\documentclass[runningheads]{llncs}
\usepackage{eccv}

\usepackage{eccvabbrv}
\usepackage{graphicx}
\usepackage{booktabs}
\usepackage[accsupp]{axessibility}

\usepackage{tikz}
\usetikzlibrary{3d}

\usepackage{pgfplots}
\pgfplotsset{compat=1.18}

\usepackage{xcolor}
\usepackage{multirow}
\usepackage{bbm}
\usepackage{amsmath}
\usepackage{amssymb}
\usepackage{subcaption}
\usepackage{caption}
\usepackage{tabularx}
\usepackage{array}
\usepackage{float}
\usepackage{siunitx}
\usepackage{pifont}
\usepackage{marvosym}
\usepackage[table]{xcolor}

\usepackage{hyperref}

\usepackage{orcidlink}

\begin{document}

\title{TRINITY: A Multi-Perspective Benchmark for Personal-Style Video Highlight Detection}
\titlerunning{TRINITY: Multi-Perspective Video Highlight Detection}

\author{
Qianqian Chen$^{*}$
\and
Hyun Bin Kim$^{*}$
\and
Denzel Elden Wijaya
\and
Yang Yi
\and
Bo Liu$^{\text{\Letter}}$
\and
Yangkai Ding$^{\text{\Letter}}$
}

\authorrunning{Q.~Chen et al.}

\institute{
Huawei Technologies Co., Ltd.
}

\maketitle

\begingroup
\renewcommand{\thefootnote}{*}
\footnotetext{Equal contribution.}
\renewcommand{\thefootnote}{\Letter}
\footnotetext{Corresponding authors.}
\endgroup

\begin{abstract}


Traditional video highlight detection relies on a narrow, event-centric definition of saliency, which often fails to generalize to unconstrained personal videos where highlights are heterogeneous and perspective-dependent. To address this, we introduce TRINITY, a multi-perspective benchmark that decomposes highlight saliency into three complementary dimensions, Event, Emotion, and Nature, within a unified temporal framework. Leveraging this multi-faceted view, we propose a shared-backbone multi-branch architecture designed for parallel multi-perspective prediction via view-specific experts. Comprehensive experiments demonstrate that our method significantly outperforms state-of-the-art baselines, achieving gains of $+7.15/+3.62$ $\text{mAP}_{\rho=15\%/50\%}$ on Mr.~HiSum and $+10.82$ mAP on YouTube Highlights. These results validate that multi-perspective modeling provides a more robust and comprehensive formulation of video saliency, especially for complex real-world scenarios. The benchmark and relevant codes will be released upon acceptance. The benchmark is available at \url{https://huggingface.co/datasets/vanilladucky/TRINITY} and the code is available at \url{https://github.com/vanilladucky/TRINITY}.


\end{abstract}

\section{Introduction}
\label{sec:intro}

Video highlight detection aims to identify moments that strongly shape human attention and memory~\cite{kahneman1993more}. Despite rapid progress, existing benchmarks largely adopt a scenario-bound and narrowly defined notion of saliency. Most datasets are constructed around sports competitions, TV programs, or query-conditioned retrieval tasks~\cite{song2015tvsum, sun2014ranking, sul2023mr, lei2021detecting}, where highlights are equated with salient semantic events or narrative peaks. 
While this single-perspective formulation has driven strong performance within controlled settings, it inherently constrains the conceptual scope of what constitutes a highlight. Consequently, models trained under this paradigm often struggle to generalize to unconstrained videos, such as those capturing daily life, where saliency is subtle, heterogeneous, and frequently decoupled from dramatic or high-intensity actions.

\begin{figure}[t] 
  \centering
  \includegraphics[width=\linewidth]{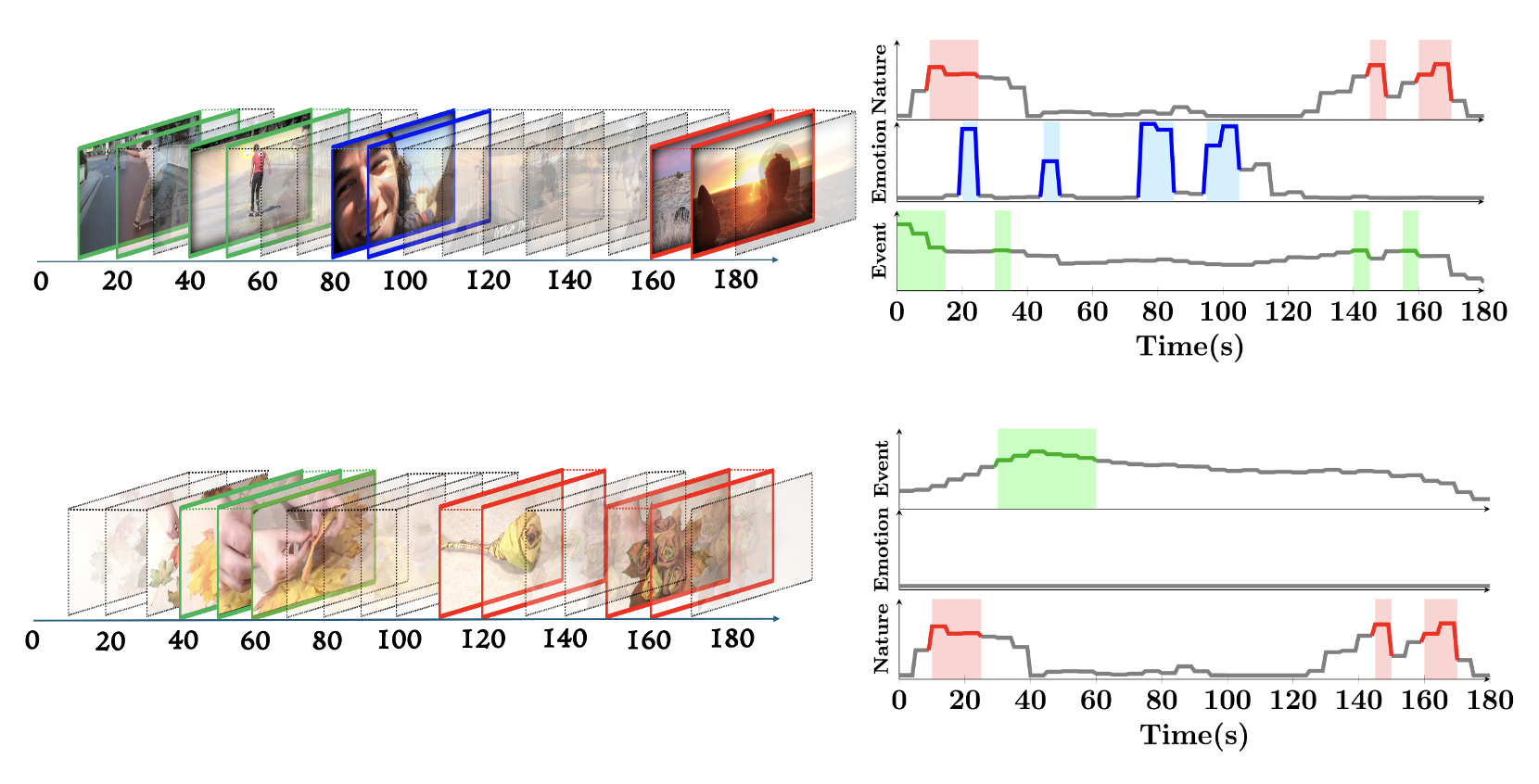}
  \caption{
Visualization of highlight distributions from three perspectives—\textcolor{green}{\emph{Event}}, \textcolor{blue}{\emph{Emotion}}, and \textcolor{red}{\emph{Nature}}—for two example videos. The horizontal axis denotes time (in seconds), and the vertical axis represents the highlight score.
  } 
  \label{fig:intro}
\end{figure}
In real life, personal-style videos dominate social media. 
These recordings typically lack explicit story structures or professionally edited climaxes~\cite{grauman2022ego4d, sellen2007life,kirk2007understanding}. 
Engaging moments may instead emerge from affective facial expressions, intimate social interactions, or aesthetically pleasing scenery, where salient signals are often multi-causal and perspective-dependent rather than single one.
For instance, in a personal vlog, highlights may stem from both scenic views and the vlogger’s joyful expressions.
Modeling highlight detection as a single-dimensional prediction task thus reduces a fundamentally multi-faceted phenomenon to a narrow formulation. 
We argue that advancing highlight detection requires moving beyond scenario-specific supervision toward a paradigm that explicitly models heterogeneous saliency mechanisms within a single video.


To this end, we introduce \textbf{TRINITY}, a multi-perspective benchmark that decomposes highlight saliency into three complementary dimensions within a unified temporal framework. 
Specifically, highlights are annotated along three perspectives: event-driven semantic peaks (denoted as \emph{Event}), affective facial moments (\emph{Emotion}), and aesthetic scenic highlights (\emph{Nature}), which may co-exist within a single video as shown in Fig.~\ref{fig:intro}.
The Emotion and Nature perspectives are constructed through a scalable automatic annotation pipeline to ensure consistency and reproducibility, while the Event perspective follows established event-centric annotations.
Importantly, the three perspectives are intentionally designed to be complementary rather than redundant, giving rise to cross-dimensional orthogonality and natural view sparsity in highlight annotations.
By disentangling these heterogeneous saliency signals, TRINITY enables a more structured analysis of highlight patterns beyond the traditional paradigm. 
Building on TRINITY, we further propose a novel shared-backbone multi-branch architecture with view-specific experts.
The shared backbone models global temporal structure and contextual coherence, while each branch specializes in localizing video highlights under a specific perspective (i.e., Event, Emotion, and Nature). 
This structural factorization enables efficient parallel multi-perspective prediction from a single input sequence, naturally aligning with the heterogeneous saliency mechanisms in personal videos. Comprehensive experiments demonstrate that our model establishes a strong baseline on TRINITY and achieves state-of-the-art performance on existing event-centric benchmarks and an emotion-centric benchmark. Extensive ablation studies further verify the effectiveness of the proposed components.

In summary, our main contributions are as follows:
\begin{itemize}




\item We introduce TRINITY, a novel multi-perspective benchmark that shifts the highlight detection paradigm from single dimension to heterogeneous saliency modeling. It decomposes highlights into three dimensions, Event, Emotion, and Nature, within a unified temporal framework.

\item We propose a shared-backbone multi-branch architecture featuring view-specific experts. This design explicitly factorizes global temporal context from perspective-specific localization, enabling the model to parallelize the detection of diverse saliency signals.

\item Comprehensive evaluations demonstrate that our approach not only establishes a robust baseline on TRINITY but also achieves state-of-the-art performance on traditional event-centric and emotion-centric benchmarks, proving its superior generalization across diverse video types.

\end{itemize}


\section{Related Work}
\label{sec:related}

\subsection{Video Highlight Detection}

Video highlight detection aims to automatically localize salient segments in untrimmed videos. 
Existing methods generally follow two main paradigms in the literature: \emph{query-conditioned} and \emph{text-agnostic} modeling. 

\paragraph{Query-conditioned highlight detection:}
QVHighlights~\cite{lei2021detecting} formulates highlight detection as query-conditioned moment retrieval with clip-level saliency prediction. Subsequent works strengthen cross-modal interaction through transformer-based grounding frameworks for query-guided localization~\cite{moon2023query, jang2023knowing, sun2024tr, liu2022umt, lin2023univtg, zeng2026promptemo}. Although such methods reduce semantic ambiguity by anchoring saliency to user intent, they rely on textual queries at inference time and are therefore unsuitable for automatic highlight detection in personal videos.

\paragraph{Text-agnostic highlight detection:} 
Text-agnostic approaches predict saliency directly from visual and audio cues. Early methods rely on edited-video supervision or ranking objectives~\cite{sun2014ranking, yao2016highlight}, while more recent works emphasize cross-category generalization and audio--visual modeling~\cite{xu2021cross, badamdorj2021joint, ye2021temporal}. Despite these advances, most approaches implicitly optimize a single notion of saliency, typically event-driven, and do not explicitly model heterogeneous highlight patterns within a unified framework. In this work, we decompose highlight saliency into complementary visual perspectives under a shared temporal protocol, enabling modeling of multiple heterogeneous patterns without textual conditioning.

\subsection{Video Highlight Detection Benchmarks}

Highlight detection is inherently subjective, and most benchmarks rely on multiple annotators to stabilize labels. Early datasets such as YouTube Highlights\cite{sun2014ranking} provide segment-level annotations, while video summarization datasets such as TVSum \cite{song2015tvsum} are often repurposed by aggregating shot-level importance scores into highlight labels. Another line of work exploits user-generated artifacts as implicit supervision. Video2GIF~\cite{gygli2016video2gif} treats user-created GIFs as highlight signals, and PHD$^2$~\cite{garcia2018phd} extends this setting to personalized highlight detection using user histories. More recently, large-scale implicit-feedback datasets such as Mr.~HiSum~\cite{sul2023mr} aggregate engagement statistics to derive highlight saliency from massive viewer interactions. Despite differences in supervision sources, most existing benchmarks implicitly assume a single notion of saliency per video. 
In contrast, our dataset decomposes highlight saliency into three complementary perspectives within a unified temporal framework, enabling structured analysis of heterogeneous highlight patterns across diverse personal videos.

\section{TRINITY Dataset}
\label{sec:dataset}

\subsection{Dataset Overview}
\label{subsec:dataset_tax}

TRINITY is a multi-perspective highlight benchmark built on open-sourced videos, organizing highlights into three perspectives: Event, Emotion, and Nature. 
Emotion- and nature-driven highlights are newly annotated over the entire video pool using a scalable two-stage pipeline. 
For the Event dimension, we directly adopt the replay-based annotations from Mr.~HiSum\cite{sul2023mr}. The following subsections describe the construction process for each perspective in detail.

\subsection{Event Perspective: Semantic Progression}
\label{subsec:ds_event}

\paragraph{Highlight Definition.}
Following the replay-based supervision mechanism introduced in Mr.\ HiSum, replay frequency derived from aggregated ``Most Replayed'' statistics is treated as a proxy for event-driven highlights. 
Segments with consistently high replay intensity are assumed to correspond to temporally localized semantic events that attract collective viewer attention.

\paragraph{Label Construction.}
Following prior practice in~\cite{Kim2025summdiff}, frame-level importance scores are aggregated onto a unified 5-second clip grid via average pooling.
To prevent peak dilution, clips containing maximal frame-level values retain the peak score. 
This preserves the relative ordering of segments.

\begin{figure}[t]
    \centering
    \begin{subfigure}[t]{0.48\linewidth}
        \centering
        \includegraphics[width=\linewidth]{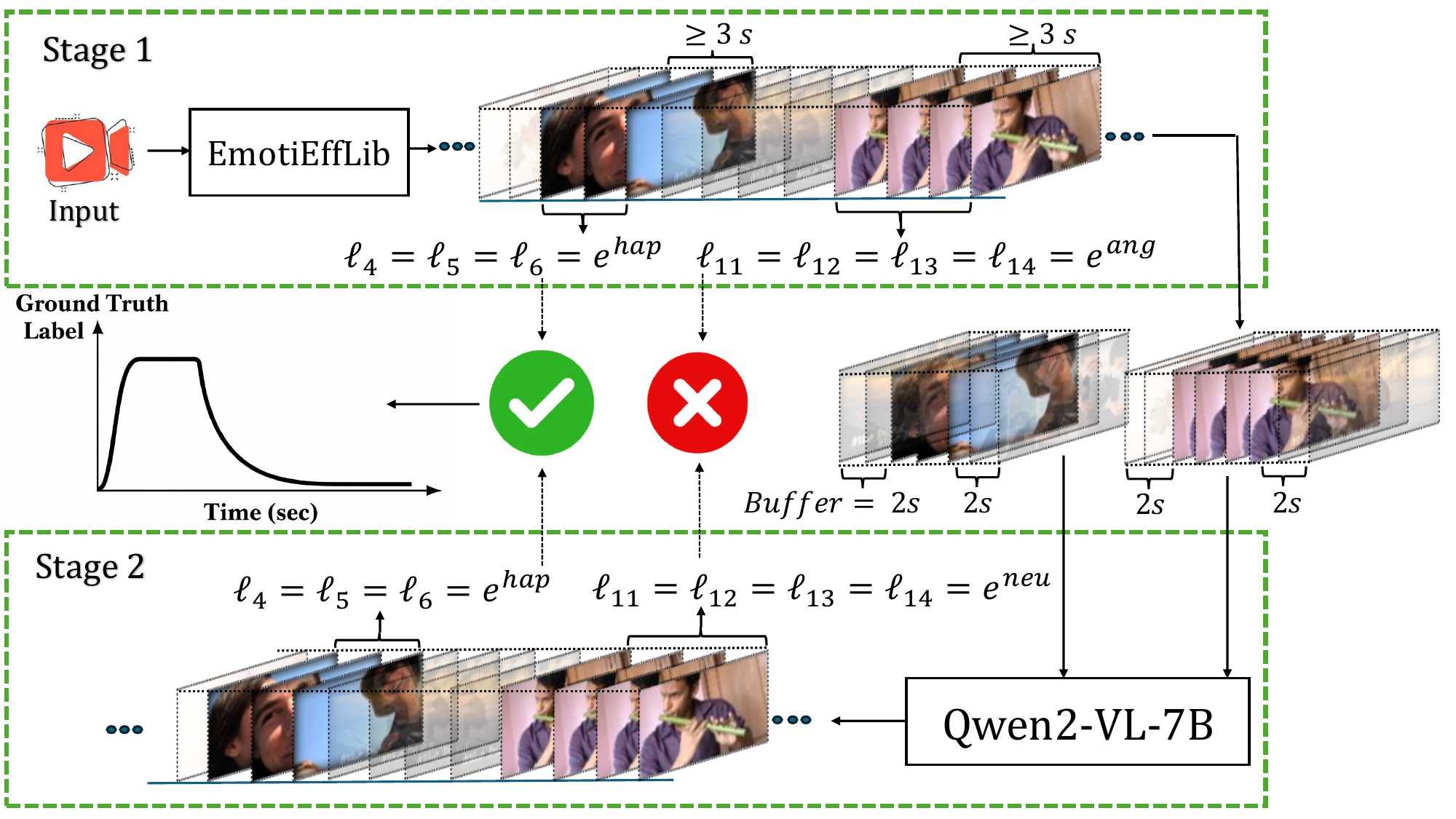}
        \caption{Two-stage pipeline for emotion-driven highlight annotation.}
        \label{fig:event_example}
    \end{subfigure}
    \hfill
    \begin{subfigure}[t]{0.48\linewidth}
        \centering
        \includegraphics[width=\linewidth]{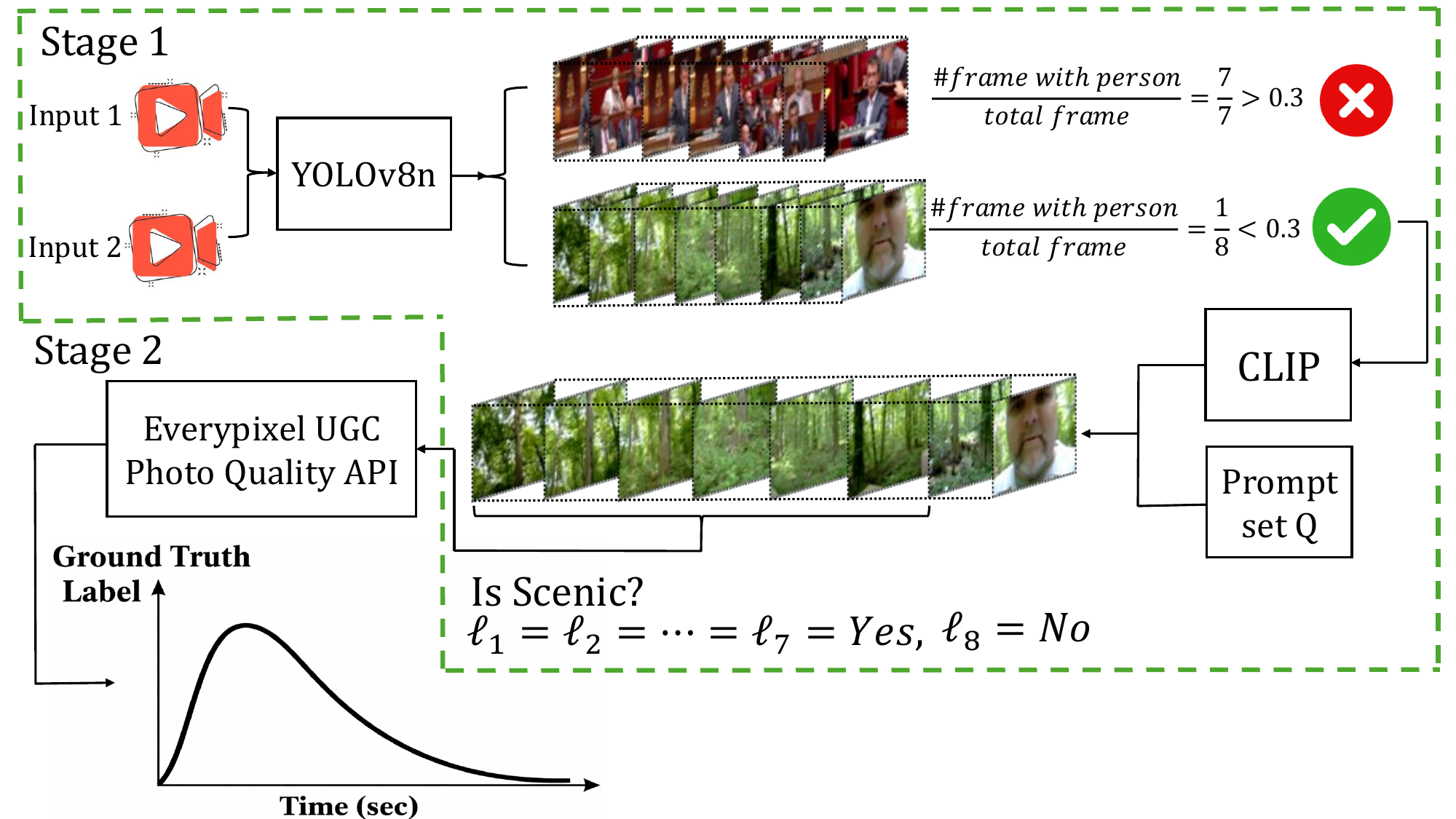}
        \caption{Two-stage pipeline for nature-driven highlight annotation.}
        \label{fig:event_random}
    \end{subfigure}
    \caption{
Overview of the annotation pipelines used to construct the TRINITY dataset.
The Emotion pipeline detects temporally stable facial expressions and verifies candidate segments via multimodal reasoning.
The Nature pipeline identifies scenic moments through semantic filtering and aesthetic assessment, retaining segments with both landscape relevance and strong visual appeal.
}
    \label{fig:trinity_event_examples}
\end{figure}

\subsection{Emotion Perspective: Affective Facial Moments}
\label{subsec:ds_emotion}

\paragraph{Highlight Definition.}
Emotion-driven highlights are segments where affective signals are conveyed through observable facial expressions and emotional reactions.

\paragraph{Label Construction.}
Emotion labels are constructed via a two-stage automatic pipeline that combines perceptual-level expression screening with independent semantic-level affect reasoning, ensuring consistent category assignments across models, as shown in Fig.~\ref{fig:trinity_event_examples} (a).

\textbf{Stage-1: Temporally consistent expression screening.}
Videos are uniformly sampled at 1\,fps to obtain a sequence of frames $\{x_t\}_{t=1}^T$.
We apply EmotiEffLib\cite{savchenko2023facial,savchenko2022classifying}, a frame-level facial expression recognition model, 
to assign labels $\ell_t \in \mathcal{E}_{1}$, where 
$\mathcal{E}_{1} =
\{e^{\text{neu}}, e^{\text{hap}}, e^{\text{sad}},
e^{\text{ang}}, e^{\text{fear}}, e^{\text{dis}}, e^{\text{sup}}\}$ 
denotes the seven basic emotion categories.
To suppress noise from transient micro-expressions, the frame $x_t$ is considered affective-valid if it belongs to a stable non-neutral window, 
that is, $\ell_t = \ell_{t+1} = \ell_{t+2} = c$ with 
$c \in \mathcal{E}_{1} \setminus \{e^{\text{neu}}\}$, 
thereby requiring emotion labels to remain stable across at least three consecutive frames. Candidate segments $\mathcal{S}$ are then constructed by merging maximal contiguous intervals of affective-valid frames. 
Each segment $s \in \mathcal{S}$ is defined as 
$s = \{ t \mid t_{\text{start}} \le t \le t_{\text{end}} \}$, 
where all frames share the same sustained non-neutral emotion category 
$c_s \in \mathcal{E}_{1} \setminus \{e^{\text{neu}}\}$.

\textbf{Stage-2: Independent context-aware multimodal screening.}
For each candidate temporal segment $s$, we construct an expanded temporal window 
$\mathcal{B}(s) = \{ t \mid t_{\text{start}} - \delta \le t \le t_{\text{end}} + \delta \}$ with $\delta = 2$. 
The corresponding clip is processed by a large vision-language model (Qwen2-VL-7B\cite{wang2024qwen2} we used), 
denoted as $\mathcal{M}(\cdot)$. It analyzes the visual and contextual cues within the expanded window and predicts an emotion category 
$\mathcal{M}(\mathcal{B}(s)) \in \mathcal{E}_{1}$.
A segment $s$ is retained as a ground-truth Emotion highlight only when
$\mathcal{M}(\mathcal{B}(s)) = c_s$, 
where $c_s$ denotes the Stage-1 emotion label.
For a retained segment $s$, we assign $y_t = 1$ for all $t \in s$, and $y_t = 0$ otherwise.

Conceptually, Stage-1 detects salient facial dynamics and expression intensity at the perceptual level. Stage-2 complements by performing semantic reasoning over multimodal context, assessing whether the segment conveys a coherent affective meaning. By keeping only segments where both models predict the same category, the pipeline yields more reliable emotion-driven highlight annotations.

\paragraph{Annotation Quality.} Annotation consistency is evaluated via cross-annotator agreement among three independent annotators on a randomly sampled set of 50 annotated segments. The inter-annotator agreement reaches 92\%, indicating stable labeling consistency across annotators. 
Under majority voting, 94\% of the emotion labels are verified as correct.
Detailed agreement analyses are reported in the Appendix.

\subsection{Nature Perspective: Scenic Aesthetic Saliency}
\label{subsec:ds_nature}

\paragraph{Highlight Definition.}
Nature-driven highlights are defined as segments dominated by visually salient scenic compositions with strong aesthetic intensity, independent of explicit human actions.

\paragraph{Label Construction.}
Nature annotations are constructed via coarse scenic localization followed by frame-level aesthetic scoring within scenic segments, enabling identification of visually appealing landscape moments (Fig.~\ref{fig:trinity_event_examples}(b)).

\textbf{Stage-1: Coarse scenic localization.}
Videos are uniformly sampled at 1\,fps to obtain frame sequences $\{x_t\}_{t=1}^T$. 
For each frame, YOLOv8n\cite{yolov8_ultralytics} is applied to detect persons, producing a binary indicator $h_t$. 
Videos whose person-frame ratio 
$R_{\text{per}} = \frac{1}{T}\sum_{t=1}^T h_t$ 
satisfies $R_{\text{per}} > \theta_v^{\text{per}}$ with $\theta_v^{\text{per}} = 0.3$ are discarded.

For the remaining videos, CLIP \cite{radford2021learning} similarities are computed between each frame $x_t$ and a curated prompt set $\mathcal{Q}$ containing landscape and human descriptions. 
Frame-level probabilities are obtained via softmax over prompt similarities and aggregated into landscape and person scores $p_t^{\text{land}}$ and $p_t^{\text{per}}$, respectively. 
A frame is marked scenic if $l_t = \mathbbm{1}\!\left[p_t^{\text{land}} > 0.4 \land p_t^{\text{land}} > p_t^{\text{per}}\right]$, indicating that the frame is more likely to depict landscape content than human-centric scenes.
All thresholds are selected based on empirical validation experiments.

\textbf{Stage-2: Frame-level aesthetic scoring.}
For each validated scenic segment $s \in \mathcal{S}$, we obtain frame-level aesthetic scores 
$\{E_t \mid t \in s, E_t \in [0,1]\}$ using the external Everypixel UGC Photo Quality, a pretrained aesthetic assessment service.  The model analyzes multiple visual attributes, including sharpness, exposure, and composition, producing a normalized aesthetic score that reflects overall perceptual quality. 
These scores are directly used as regression targets for the Nature dimension. 
Specifically, the ground-truth label for a frame $t$ is defined as 
$y_t = E_t$ if $\exists s \in \mathcal{S}$ such that $t \in s$, and $y_t = 0$ otherwise. This stage is applied only when $\mathcal{S} \neq \emptyset$, ensuring supervision is derived from regions that satisfy both scenic alignment and strong aesthetic quality.

\paragraph{Annotation Quality.}
Given the subjective nature of aesthetic judgment, we conduct a human audit with three independent annotators on 50 randomly sampled videos. 
For scenic validity, annotators label scenic frames for each video, and consensus scenic segments are derived via majority voting over frame-level labels. 
The average IoU between the consensus segments and the labeled ground-truth segments is 0.89727, while the average pairwise inter-annotator IoU is 0.89744, 
indicating that the dataset labels align with human annotations at a level nearly identical to the agreement among annotators. 
For aesthetic quality validation, 94\% of the aesthetic highlights are validated by majority voting over the annotators’ judgments, with an inter-annotator agreement rate of 90\%. To further ensure reproducibility and mitigate proprietary-model bias, we evaluate the cross-scorer agreement of our Everypixel-based\footnote{\url{https://api.everypixel.com/v1/quality_ugc}} Nature labels against alternative open-source aesthetic models (MUSIQ \cite{Ke2021MUSIQ} and LAION-Aesthetics\footnote{\url{https://github.com/christophschuhmann/improved-aesthetic-predictor}}) across 5,651 annotated videos. Our labels exhibit strong alignment, achieving an IoU of 0.665 at $\rho=15\%$ and 0.956 at $\rho=50\%$ with MUSIQ, and 0.661 at $\rho=15\%$ and 0.955 at $\rho=50\%$ with LAION, confirming that the annotations capture general, scorer-agnostic aesthetic signals.

\subsection{Statistical Overview}
\label{subsec:statistical_overview}

To provide a comprehensive view of the proposed benchmark, Table~\ref{tab:trinity_perspective_stats} summarizes the global statistics of the three highlight perspectives across the entire TRINITY dataset, including the total number of videos, annotated highlight peaks, and the average duration per peak.

\begin{table}[tb]
\centering
\caption{
Global statistics of the three highlight perspectives in the full TRINITY dataset.
}
\label{tab:trinity_perspective_stats}
\small
\setlength{\tabcolsep}{10pt}
\renewcommand{\arraystretch}{1.12}
\resizebox{\linewidth}{!}{
\begin{tabular}{l c c c}
\toprule
\textbf{Perspective} & \textbf{\#Videos} & \textbf{\#Peaks} & \textbf{Avg. Peak Dur. (s)} \\
\midrule
TRINITY-Nature  & 15,540 & 15,540 & 10.34 \\
TRINITY-Emotion & 16,257 & 32,046 & 4.87 \\
TRINITY-Event   & 27,845 & 55,649 & 9.90 \\
\bottomrule
\end{tabular}
}
\end{table}

The statistics reveal clear domain and temporal distinctiveness across the three formulated perspectives. The TRINITY-Event perspective comprises the largest video pool and peak count, capturing broad semantic activities. In contrast, the temporal characteristics vary significantly between Emotion and Nature: emotion-driven highlights are typically short and highly concentrated ($\sim$4.87\,s), capturing transient facial and behavioral expressions; conversely, nature-driven highlights feature longer sustained durations ($\sim$10.34\,s), as aesthetic scenic compositions generally unfold over continuous camera periods. 

These macro-level structural variations inherently support our core motivation to model video highlights via a multi-perspective formulation. To further investigate how these perspectives interact and validate their statistical orthogonality, we provide a detailed cross-perspective temporal overlap analysis on the co-annotated subset in the Appendix.

\section{Methodology}
\label{sec:meth}

\begin{figure*}[t]
  \centering
  \includegraphics[width=\linewidth]{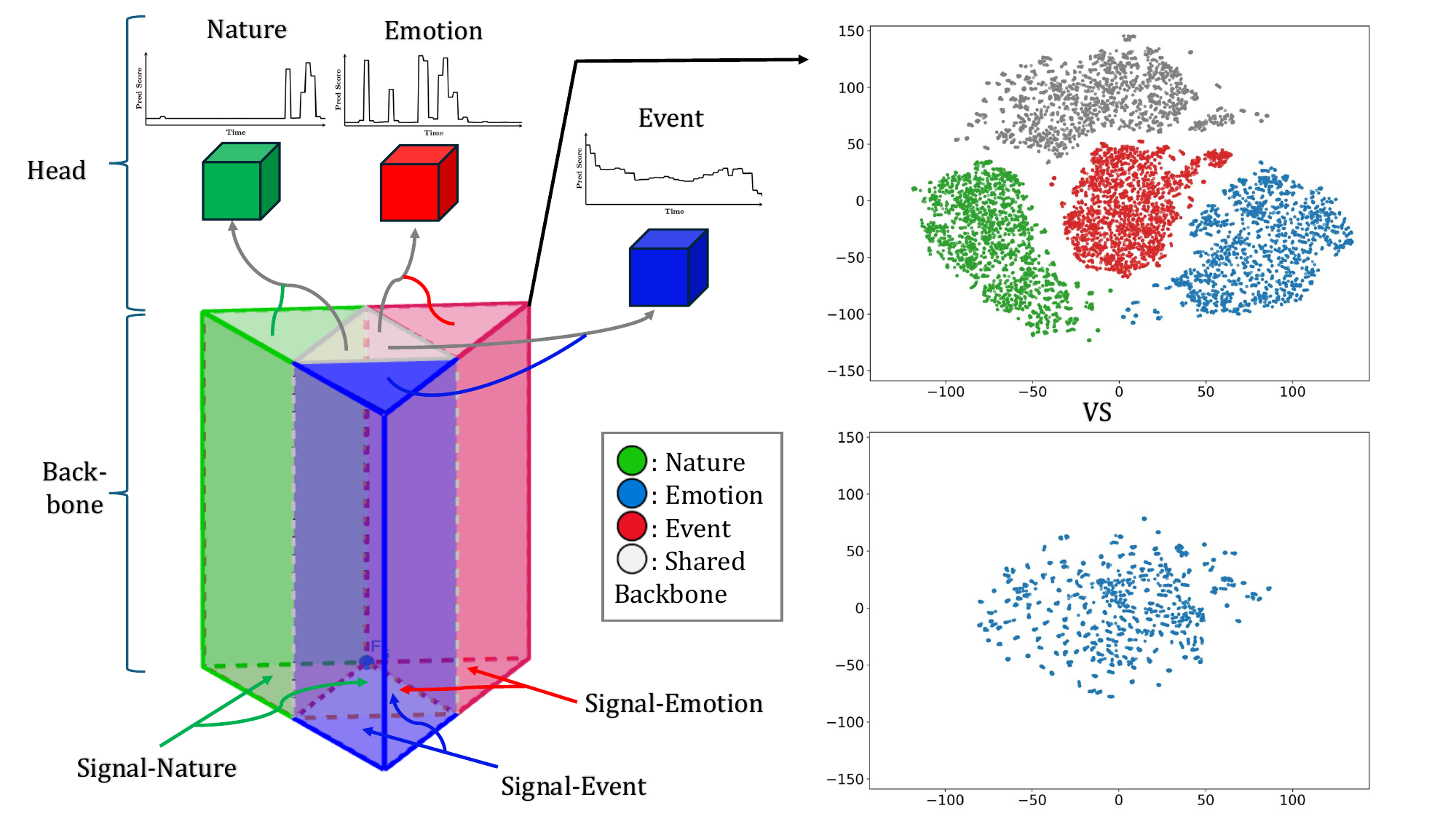}
  \caption{Left. Overview of the proposed architecture with a shared backbone and perspective-specific experts. 
Right. Visualization of the learned embedding space, where private experts form distinct clusters while the shared backbone captures global temporal context. 
Compared with a single-task event-only model (bottom-right), the proposed design yields a structured and separable representation space across perspectives.}
  \label{fig:main_results}
\end{figure*}

\subsection{Overall Architecture}
As illustrated in \cref{fig:main_results}, we formulate video highlight detection as a \emph{multi-task} temporal segment scoring problem with three highlight perspectives: \textbf{Event} ($t{=}e$), \textbf{Emotion} ($t{=}m$), and \textbf{Nature} ($t{=}n$).
Given a video $v$, we split it into $N$ temporal segments $\{s_i\}_{i=1}^{N}$, where each segment represents a non-overlapping 5-second interval. The emotion task ($t=m$) provides a binary highlight label $y_i^{t}\in\{0,1\}$, reflecting the discrete presence of facial expressions, while the event ($t=e$) and nature ($t=n$) tasks provide continuous labels $y_i^{t}\in[0,1]$ derived from replay statistics and aesthetic scores, respectively. Each segment is represented by a pre-extracted 512-dimensional feature vector $x_i \in \mathbb{R}^{512}$.


Our architecture contains one \emph{shared} temporal backbone $T_s$ and three \emph{private} temporal backbones $\{T_p^{t}\}_{t\in\{e,m,n\}}$. All backbones are instantiated as Transformer encoders. In contrast to permutation-invariant set encoders, we explicitly model \emph{ordered} temporal segments and inject relative temporal information via 1D RoPE \cite{su2024roformer} within self-attention layers.

The shared backbone $T_s$ is designed to capture task-agnostic, global temporal context across video timestamps. In parallel, each private backbone $T_p^{t}$ specializes in task-specific temporal patterns, emphasizing sharp and sparse temporal activations that are characteristic of task $t$. Let $X=[x_1,\dots,x_N]$ denote the sequence of segment embeddings. 
Concretely, for each task $t$, we parameterize the shared and private streams with distinct query, key, and value projections, i.e.,
\begin{equation}
\begin{aligned}
Q_s &= X W_{Q}^{(s)}, \quad &K_s &= X W_{K}^{(s)}, \quad &V_s &= X W_{V}^{(s)},\\
Q_p^{t} &= X W_{Q}^{(p,t)}, \quad &K_p^{t} &= X W_{K}^{(p,t)}, \quad &V_p^{t} &= X W_{V}^{(p,t)} .
\end{aligned}
\end{equation}
These separate projection matrices encourage the shared and task-private representations to occupy different regions of the latent space, enabling complementary modeling of global context and task-specific temporal cues.

The input sequence $X$ is passed through the shared backbone to produce a globally contextualized embedding, while the same sequence is processed by the task-specific private backbone to produce a task-specific embedding. These embeddings are concatenated to capture both global and task-specific semantics. Finally, three independent prediction heads $\{C^t\}$ (one per task) are used, each mapping $z_i^{t}$ to a highlight probability. Each prediction head consists of multiple linear layers with ReLU activations. 

\begin{equation}
\hat{y}_i^{t} = \sigma \left( C^{t} [T_s(x_i), T_{p}^{t}(x_i)] \right), \quad \text{where } \sigma \text{ is the sigmoid function.}
\end{equation}

\subsection{Task-Conditioned Expert Routing}
Unlike soft MoE gating, our routing is \emph{deterministic} and purely task-conditioned.
During training, we maintain three different dataloaders (Event/Emotion/Nature). A sample drawn from the dataloader for task $t$ is routed to both the corresponding private backbone $T_p^{t}$ and the shared backbone $T_s$.

In each iteration, we process three mini-batches at a time (one per dataloader) in a round robin fashion, yielding three different task losses and their gradients. The losses are backpropagated only after the three mini-batches are passed through. This design ensures that: (i) each private backbone only specializes in its own supervision, (ii) the shared backbone is trained in all tasks, and (iii) any possible catastrophic forgetting \cite{McCloskey1989} between datasets is avoided.

\subsection{Total Loss}
We adopt a binary cross-entropy (BCE) objective for all tasks. For a mini-batch of size $B$ and sequence length $N$, the task loss is
\begin{equation}
\mathcal{L}_t
= \frac{1}{BN}\sum_{b=1}^{B}\sum_{i=1}^{N}
\mathrm{BCE}\!\left(y_{b,i}^{t}, \hat{y}_{b,i}^{t}\right),
\qquad t\in\{e,m,n\}.
\end{equation}
The model is then trained using the sum of these three individual task losses with equal weights.
\begin{equation}
\mathcal{L}_{total} = \sum_{t \in \{e, m, n\}} \lambda_t \mathcal{L}_t, \quad \text{where} \ \lambda_e = \lambda_m = \lambda_n = 1.
\end{equation}

\subsection{Inference Strategy}
During inference, the segment embeddings of a video are fed to the model and processed by the shared backbone as well as each of the three private backbones. Subsequently, for each task, the shared backbone output and the corresponding private backbone output are fused and passed through the corresponding prediction head. As a result, multi-perspective highlight scores $\hat{y}_i^{e}, \hat{y}_i^{m}, \hat{y}_i^{n}$ are produced for each temporal segment.

\section{Experiments}
\label{sec:exp}

\subsection{Datasets and Metrics}
\paragraph{Pretraining.}
We pretrain on TRINITY, a unified multi-perspective highlight dataset consisting of three perspective groups: Event, Emotion, and Nature. 
These groups are derived by grouping videos according to highlight perspectives. 
TRINITY-Event contains \textbf{27,845} videos, TRINITY-Emotion contains \textbf{16,257} videos, and TRINITY-Nature contains \textbf{15,540} videos, corresponding to event-, emotion-, and nature-centric annotations, respectively. 
We adopt a joint multi-dataset training strategy across the three perspectives. 
Each video is segmented into fixed-length 5-second temporal segments following prior practice in~\cite{Kim2025summdiff}. 
Each segment is assigned a normalized highlight score within the range $[0,1]$, reflecting the degree of saliency under the corresponding perspective.

\paragraph{Benchmarks and Evaluation Metrics.}
We assess the event-centric capability of our model by fine-tuning only the Event branch while freezing all other components. Performance is evaluated on YouTube Highlights~\cite{sun2014ranking} and Mr.~HiSum~\cite{sul2023mr}. 
Crucially, we adopt the official train/validation/test splits of Mr.~HiSum to ensure fair comparability with prior work. 
Since part of our pretraining data is sourced from Mr.~HiSum, all test videos are removed from the pretraining pool to prevent data leakage. 
Mr.~HiSum is released under the CC BY 4.0 license, allowing modifications; to adhere to this, we will distribute only high-level features and our new annotations.
For Mr.~HiSum and TRINITY, we report $\mathrm{mAP}_{\rho=50\%}$ and $\mathrm{mAP}_{\rho=15\%}$, and for YouTube Highlights, we report the standard mAP metric. 
For emotion modeling, we fine-tune only the Emotion branch and freeze the remaining modules. 
We evaluate on VEATIC~\cite{ren2024veatic}, a context-aware video emotion dataset with continuous valence and arousal annotations. 
Its character-centric and temporal affective supervision aligns well with our Emotion perspective. 
Crucially, the Emotion branch focuses on emotion-driven highlight localization under skewed web distributions rather than balanced recognition, using VEATIC as auxiliary evidence for temporally grounded affective cues. While common affective patterns are well-supported, addressing rare emotions and direct external validation remains future work.
We report SAGR and RMSE following the official protocol.

\subsection{Implementation Details}
Our model is implemented in PyTorch. Segment-level features are extracted using a frozen CLIP ViT-B/32 encoder. Temporal backbones are Transformer encoder layers with hidden size $512$, $8$ attention heads, and depth $5$, using 1D RoPE for temporal position encoding. For each perspective, the forward pass consists of $4$ shared attention heads and $4$ task-specific private heads, yielding $8$ heads per branch. During TRINITY pretraining, one shared backbone and three perspective-specific private backbones (Event, Emotion, Nature) are jointly optimized with three task heads. Optimization uses AdamW, with early stopping if validation performance does not improve for more than three consecutive epochs across all three tasks. The learning rate, weight decay, batch size, and dropout ratio are set to $5e^{-5}$, $0.01$, $16$, and $0.2$, respectively.

\begin{table*}[t]
\centering
\caption{
Fine-tuned highlight detection results on Mr.~HiSum, YouTube Highlights, and VEATIC.
\textbf{Bold} indicates the best result and \underline{underlined} values indicate the second best.
}
\label{tab:sota_final_v2}
\setlength{\tabcolsep}{3.8pt} 
\renewcommand{\arraystretch}{1.2}
\resizebox{\textwidth}{!}{
\begin{tabular}{l|cc|lccccccc|l|cc}
\toprule
\multicolumn{3}{c|}{\textbf{Mr.~HiSum}} & 
\multicolumn{8}{c|}{\textbf{YouTube Highlights}} & 
\multicolumn{3}{c}{\textbf{VEATIC}} \\
\cmidrule(lr){1-3} \cmidrule(lr){4-11} \cmidrule(lr){12-14}
Method & $\mathrm{mAP}_{\rho=15\%}$ & $\mathrm{mAP}_{\rho=50\%}$ 
& Method & Dog & Gym & Park. & Skate. & Skii. & Surf. & Avg. 
& Method & SAGR$\uparrow$ & RMSE$\downarrow$ \\ 
\midrule

SAVS-Net\cite{alharbi2024effective}  
& - & 57.12 
& Temp.-Cue\cite{ye2021temporal} 
& 55.38 & 62.66 & 70.88 & 69.06 & 60.05 & 59.76 & 62.97 
& RB-SFP\cite{ren2024region} 
& - & 0.2423 \\

SL-Module\cite{xu2021cross}
& 24.95 & 58.63
& Mr.~HiSum\cite{sul2023mr}  
& 50.80 & 67.20 & \underline{78.30} & 42.00 & 61.80 & 71.60 & 62.00 
& VEATIC\cite{ren2024veatic} 
& 0.7637 & 0.2410 \\

PGL-SUM\cite{koutras2021combining}
& 27.45 & 61.60
& RASL\cite{li2024unsupervised}      
& 64.20 & \underline{75.80} & 70.90 & 45.60 & 66.50 & 66.70 & 65.10 
& VDFS\cite{saigo2025enhancing} 
& \underline{0.804}  & \underline{0.182} \\

HierSum\cite{beedu2025hiersum}   
& 32.60 & 63.80 
& C2\cite{zhao2025weakly}        
& - & - & - & - & - & - & 67.20 
& - & - & - \\

Aha\cite{Chang2025aha}       
& 32.66 & 64.19 
& AV-Recur.\cite{islam2025unsupervised} 
& - & - & - & - & - & - & 68.30 
& -- & -- & -- \\
        
SummDiff\cite{Kim2025summdiff}  
& \underline{33.83} & \underline{65.44} 
& SL-Module\cite{xu2021cross}   
& 70.80 & 53.20 & 77.20 & \underline{72.50} & 66.10 & 76.20 & 69.30 
& -- & -- & -- \\

--        
& -- & --    
& PLD-VHD\cite{wei2022learning}   
& \textbf{74.90} & 70.20 & 77.90 & 57.50 & \underline{70.70} & \underline{79.00} & \underline{73.00} 
& -- & -- & -- \\

\midrule
\rowcolor[HTML]{F2F2F2} 
\textbf{Ours} 
& \textbf{40.98} & \textbf{69.06} 
&  
& \underline{74.87} & \textbf{91.89} & \textbf{87.70} & \textbf{83.95} & \textbf{77.20} & \textbf{87.33} & \textbf{83.82} 
&  
& \textbf{0.8185} & \textbf{0.1749} \\

\bottomrule
\end{tabular}
}
\end{table*}

\subsection{Quantitative Results}
\paragraph{Performance on Public Benchmarks.}
We compare our method with recent state-of-the-art approaches on 
Mr.HiSum, YouTube Highlights, and VEATIC. 
Results are summarized in Table~\ref{tab:sota_final_v2}. On Mr.HiSum, our method achieves 40.98 $\mathrm{mAP}_{\rho=15\%}$ and 
69.06 $\mathrm{mAP}_{\rho=50\%}$, outperforming the recent 
SummDiff~\cite{Kim2025summdiff} by \textbf{+7.15} and \textbf{+3.62} points, respectively. 
The gains are consistent under both strict and relaxed evaluation ratios.
On YouTube Highlights, under the text-agnostic setting, we achieve the best average performance and leading results on most categories, 
reaching an average score of 83.82, 
exceeding the strongest prior baseline PLD-VHD\cite{wei2022learning} by \textbf{+10.82} points. 
Large improvements are observed in dynamic categories such as Gym, Skating, and Surfing, 
indicating strong cross-domain generalization.
On VEATIC, our model obtains 0.8185 SAGR and 0.1749 RMSE, 
surpassing previous methods on both correlation and regression precision. 
This demonstrates that our framework generalizes beyond ranking-based metrics. 

Across three heterogeneous benchmarks with distinct evaluation protocols, 
our method consistently establishes new state-of-the-art results, 
validating the effectiveness of our shared--private highlight modeling design.

\begin{table}[t]
  \centering
  \caption{
State-of-the-art comparison under the three TRINITY highlight perspectives 
(Nature, Emotion, and Event), reflecting the multi-perspective formulation of TRINITY. 
\textbf{Bold} indicates the best result.
}
  \label{tab:trinity_sota_narrow}
  \renewcommand{\arraystretch}{1.2}
  \setlength{\tabcolsep}{4.5pt} 
  \resizebox{0.85\textwidth}{!}{
    \begin{tabular}{l|cc|cc|cc}
      \toprule
      \multirow{3}{*}{\textbf{Method}} 
      & \multicolumn{2}{c|}{\textbf{TRINITY-Nature}} 
      & \multicolumn{2}{c|}{\textbf{TRINITY-Emotion}} 
      & \multicolumn{2}{c}{\textbf{TRINITY-Event}} \\
      \cmidrule(lr){2-3} \cmidrule(lr){4-5} \cmidrule(lr){6-7}
      & $\mathrm{mAP}_{\rho=15\%}$ 
      & $\mathrm{mAP}_{\rho=50\%}$ 
      & $\mathrm{mAP}_{\rho=15\%}$ 
      & $\mathrm{mAP}_{\rho=50\%}$ 
      & $\mathrm{mAP}_{\rho=15\%}$ 
      & $\mathrm{mAP}_{\rho=50\%}$ \\
      \midrule

      SL-Module \cite{xu2021cross}
      & 38.68 & 62.23
      & 50.55 & 67.57 
      & 31.46 & 62.71 \\

      PGL-SUM \cite{koutras2021combining}
      & 46.40 & 61.84
      & 62.47 & \textbf{73.81} 
      & 33.61 & 61.84 \\

      CSTA \cite{hong2024csta}
      & 36.47 & 39.67
      & 64.96 & 67.29 
      & 35.97 & 40.77 \\
      
      SumDiff \cite{Kim2025summdiff}
      & 35.68 & 59.63
      & 37.17 & 61.38
      & 26.21 & 57.45 \\
      
      Aha$^{\dagger}$ \cite{Chang2025aha}
      & 38.69 & 40.02
      & 21.02 & 28.29
      & 32.66 & 64.19 \\
      
      Qwen3-VL-8B$^{\dagger}$ \cite{Bai2025Qwen3VL}
      & 24.89 & 56.11
      & 22.99 & 53.57
      & 17.92 & 52.46 \\
      
      Qwen3-VL-235B$^{\dagger}$ \cite{Bai2025Qwen3VL}
      & 27.91 & 58.29
      & 26.42 & 56.90
      & 18.10 & 52.82 \\

      \midrule
      \rowcolor[HTML]{F2F2F2}
      \textbf{Ours} 
      & \textbf{58.74} & \textbf{68.05} 
      & \textbf{67.42} & 71.32 
      & \textbf{40.98} & \textbf{69.06} \\
      \bottomrule
    \end{tabular}
  }
  \vspace{2pt} \\
  \leftline{\scriptsize \hspace{0.125\textwidth} $^{\dagger}$ Zero-shot due to fine-tuning cost.}
\end{table}
\paragraph{Evaluation on TRINITY.}
We evaluate several representative highlight detection methods~\cite{xu2021cross, koutras2021combining, hong2024csta} 
under the three highlight perspectives defined in TRINITY (Nature, Emotion, and Event), as shown in Table~\ref{tab:trinity_sota_narrow}. 
Our framework consistently achieves the best overall performance across the three perspectives under stricter evaluation ratios. 
Under the Nature perspective, our method attains 58.74 $\mathrm{mAP}_{\rho=15\%}$ and 68.05 $\mathrm{mAP}_{\rho=50\%}$, substantially outperforming prior methods and demonstrating a stronger capability in capturing scenery-driven saliency. 
Under the Emotion perspective, we obtain the best $\mathrm{mAP}_{\rho=15\%}$ score of 67.42 while achieving competitive performance at $\mathrm{mAP}_{\rho=50\%}$ (71.32). 
Under the Event perspective, our model also achieves the best performance under both metrics, reaching 40.98 $\mathrm{mAP}_{\rho=15\%}$ and 69.06 $\mathrm{mAP}_{\rho=50\%}$.

Notably, although all methods are trained on TRINITY, baseline architectures originally designed for single-definition saliency modeling exhibit limitations when handling heterogeneous highlight signals, particularly for the Nature and Emotion perspectives. In contrast, our shared–private architecture separates perspective-specific representations while maintaining a unified temporal backbone, enabling robust highlight localization across diverse saliency definitions.

\begin{table}[t]
\centering
\caption{
Architecture ablation on TRINITY. 
We analyze the impact of shared backbones, private experts, and multi-task learning. 
Results are reported using $\mathrm{mAP}_{\rho=15\%}$ and $\mathrm{mAP}_{\rho=50\%}$ (denoted as 15\% and 50\% in the header).
}
\label{tab:ablation_shared_private}

\setlength{\tabcolsep}{3.8pt}
\renewcommand{\arraystretch}{1.2}

\resizebox{\textwidth}{!}{
\begin{tabular}{l|ccccc|cc|cc|cc|cc}
\toprule
\multirow{2}{*}{\textbf{Setting}} &
\multicolumn{5}{c|}{\textbf{Arch. Config.}} &
\multicolumn{2}{c|}{\textbf{Event}} &
\multicolumn{2}{c|}{\textbf{Nature}} &
\multicolumn{2}{c|}{\textbf{Emotion}} &
\multicolumn{2}{c}{\textbf{Avg.}} \\
\cmidrule(lr){2-6}\cmidrule(lr){7-8}\cmidrule(lr){9-10}\cmidrule(lr){11-12}\cmidrule(lr){13-14}
& \#B.B. & Sha. & Pri. & \#Head & MTL
& 15\% & 50\%
& 15\% & 50\%
& 15\% & 50\%
& 15\% & 50\% \\
\midrule

S1: Single-Task Baseline
& 1 & \checkmark & \texttimes & 1 & \texttimes
& 32.85 & 63.06
& 43.16 & 65.57
& 61.05 & 75.56
& 45.69 & 68.06 \\

S2: Shared-Backbone MTL
& 1 & \checkmark & \texttimes & 3 & \checkmark
& 37.97 & 66.86
& 44.41 & 64.35
& 65.20 & 75.71
& 49.19 & 68.97 \\

S3: Independent Experts
& 3 & \texttimes & \texttimes & 1 & \texttimes
& 37.54 & 67.51
& 44.78 & 65.26
& 63.27 & 76.34
& 48.53 & 69.70 \\

S4: Shared--Private Experts
& 4 & \checkmark & \checkmark & 3 & \checkmark
& 40.98 & 69.06
& 58.74 & 68.05
& 67.42 & 71.32
& 55.71 & 69.48 \\

\bottomrule
\end{tabular}
}
\end{table}

\begin{table}[t]
\centering
\caption{Dataset contribution ablation under a fixed architecture (S4: Shared--Private Experts).
Models are trained with progressively richer supervision.}
\label{tab:dataset_ablation}

\setlength{\tabcolsep}{3.8pt}
\renewcommand{\arraystretch}{1.2}

\resizebox{\textwidth}{!}{
\begin{tabular}{l|ll|ll|ll|ll}
\toprule
\multirow{2}{*}{\textbf{Setting}} 
& \multicolumn{2}{c|}{\textbf{Event}}
& \multicolumn{2}{c|}{\textbf{Nature}}
& \multicolumn{2}{c|}{\textbf{Emotion}}
& \multicolumn{2}{c}{\textbf{Avg.}} \\
\cmidrule(lr){2-3}\cmidrule(lr){4-5}\cmidrule(lr){6-7}\cmidrule(lr){8-9}
& $mAP_{\rho=15\%}$ & $mAP_{\rho=50\%}$
& $mAP_{\rho=15\%}$ & $mAP_{\rho=50\%}$
& $mAP_{\rho=15\%}$ & $mAP_{\rho=50\%}$
& $mAP_{\rho=15\%}$ & $mAP_{\rho=50\%}$ \\
\midrule

D1: Event
& 40.25 & 69.45
& 31.25 & 65.58
& 40.83 & 65.76
& 37.44 & 66.93 \\

D2: Event + Nature
& 39.98 {\color{red}{$\downarrow$0.27}}
& 68.70 {\color{red}{$\downarrow$0.75}}
& 57.89 {\color{ForestGreen}{$\uparrow$26.64}}
& 67.71 {\color{ForestGreen}{$\uparrow$2.13}}
& 45.98 & 61.06
& 48.25 & 64.32 \\

D3: Full (Event + Nature + Emotion)
& 40.98 {\color{ForestGreen}{$\uparrow$1.00}}
& 69.06 {\color{ForestGreen}{$\uparrow$0.36}}
& 58.74 {\color{ForestGreen}{$\uparrow$0.85}}
& 68.05 {\color{ForestGreen}{$\uparrow$0.34}}
& 67.42 {\color{ForestGreen}{$\uparrow$21.44}}
& 71.32 {\color{ForestGreen}{$\uparrow$10.26}}
& 55.71 & 69.48 \\

\bottomrule
\end{tabular}
}
\end{table}

\subsection{Ablation Study}
\label{sec:ablation}

\paragraph{Architecture ablation.}
Table~\ref{tab:ablation_shared_private} analyzes the effect of architectural decomposition. 
Moving from single-task training (S1) to a shared-backbone multi-task model (S2) 
improves the average performance by +3.50 $\mathrm{mAP}_{\rho=15\%}$, 
indicating that cross-task supervision benefits shared temporal representations. Using fully independent experts (S3) improves performance under 
$\mathrm{mAP}_{\rho=50\%}$ compared with shared-backbone MTL (S2), indicating improved ranking when broader temporal coverage is evaluated, as the relaxed threshold ($\rho=50\%$) rewards highlights that capture a larger portion of the ground-truth segments.
However, the absence of shared representations limits overall cross-perspective generalization. 
Our shared--private expert design (S4) achieves the best overall performance 
(55.71 / 69.48). Relative to the independent-expert setting (S3), 
S4 improves Event from 37.54 to 40.98, Nature from 44.78 to 58.74, and Emotion from 63.27 to 67.42 under $\mathrm{mAP}_{\rho=15\%}$, demonstrating the benefit of combining shared representations with perspective-specific experts. Consistent with this observation, as illustrated in \cref{fig:main_results}, 
the t-SNE visualization shows clearer perspective-wise clustering of temporal segment features under S4, 
indicating more structured temporal representations across perspectives.

\paragraph{Dataset contribution ablation.}
Table~\ref{tab:dataset_ablation} examines the impact of progressively incorporating multi-perspective supervision under a fixed architecture (S4).
Training with Event-only data (D1) yields moderate performance across all three dimensions. 
Adding Nature supervision (D2) substantially improves Nature performance 
from 31.25 to 57.89 under $\mathrm{mAP}_{\rho=15\%}$, 
while Event performance remains largely stable, 
indicating limited negative interference across different perspectives. 
Finally, incorporating all three perspectives (D3) provides consistent improvements across dimensions, with the largest gain on Emotion 
The full multi-perspective training achieves the highest overall average performance (55.71 / 69.48), 
validating the benefit of jointly modeling heterogeneous highlight cues. 

\subsection{Gradient Conflict Observation}
A notable problem in multi-task learning is gradient conflict between the losses of different tasks. Conflicting gradients were shown to have negative transfer effects\cite{lee2018deep} and there have been multiple attempts to mitigate this effect \cite{cha2021conflict, yu2020gradient, senushkin2023independent}. Since our problem is also formulated as a multi-task learning problem, we performed gradient conflict analysis to ensure there were no adversarial effects. In order to measure the gradient conflict, we employed cosine similarity of gradients \cite{yu2020gradient} between our three different task losses. 

\begin{figure}[t]
    \centering
    \begin{subfigure}{0.48\linewidth}
        \centering
        \includegraphics[width=\linewidth]{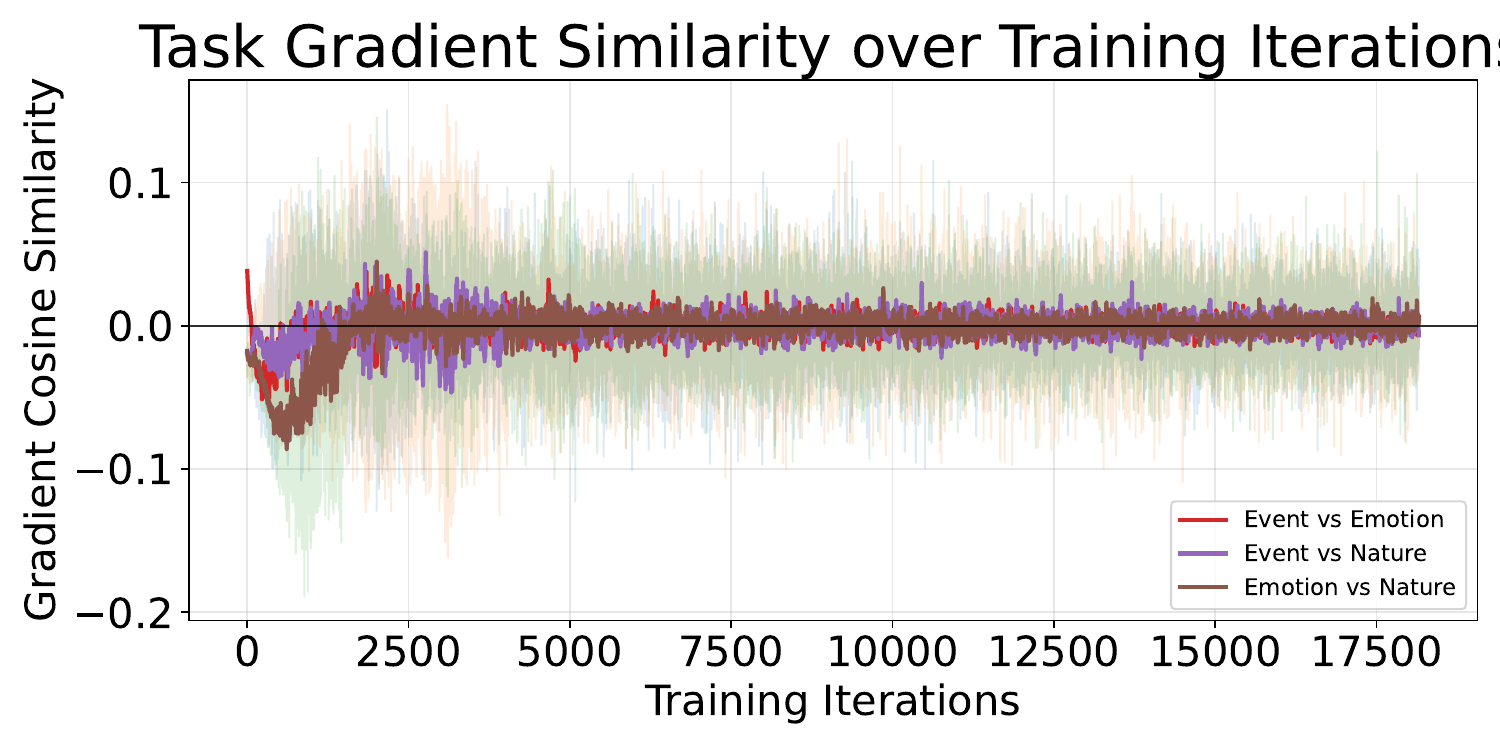}
        \caption{With Aligned-MTL}
        \label{fig:left_plot_gradient}
    \end{subfigure}
    \hfill
    \begin{subfigure}{0.48\linewidth}
        \centering
        \includegraphics[width=\linewidth]{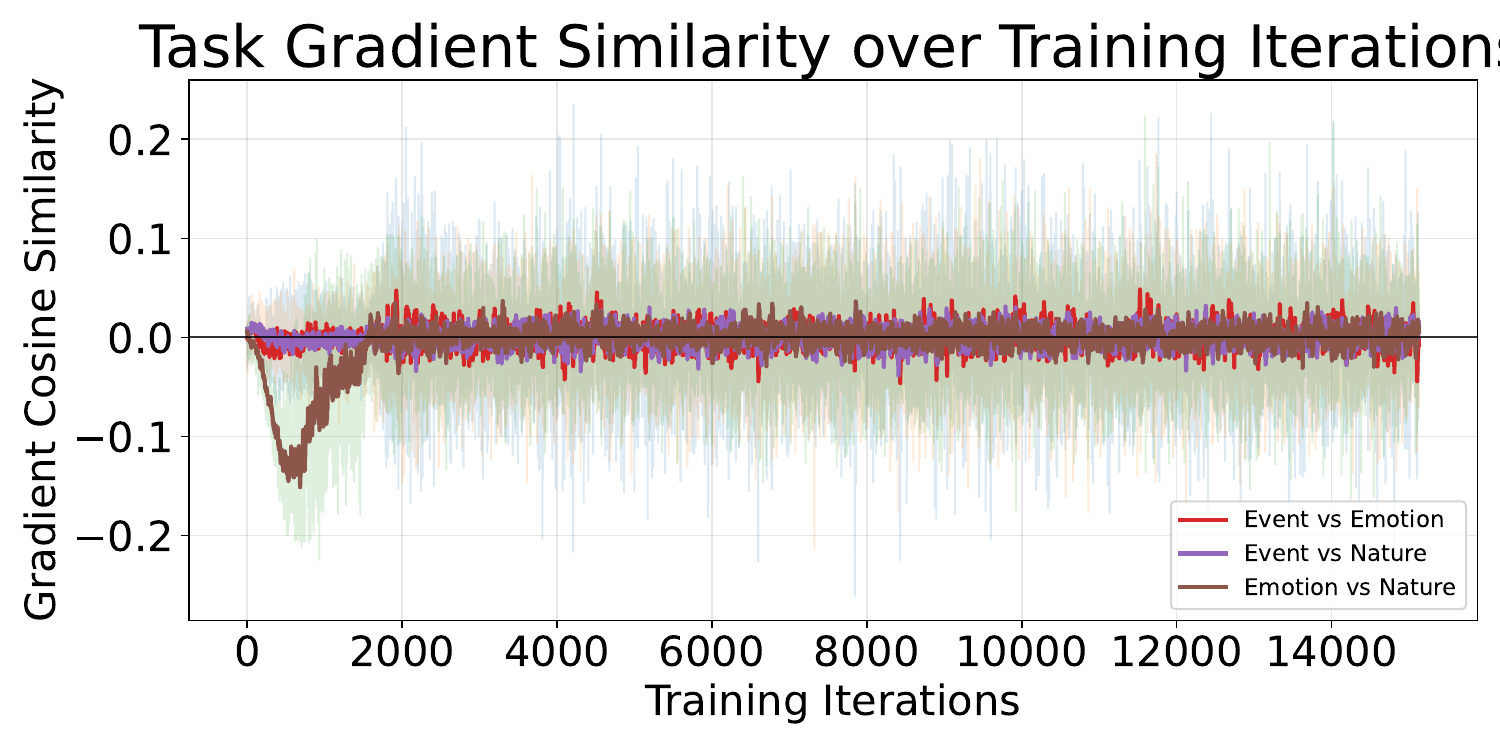}
        \caption{Without Aligned-MTL}
        \label{fig:right_plot_gradient}
    \end{subfigure}
    
    \caption{Comparison of gradient cosine similarity during training with and without Aligned-MTL.}
    \label{fig:side_by_side}
\end{figure}

The gradient cosine similarity of our model can be seen in \cref{fig:right_plot_gradient} where the $\cos(\theta_{i,j})<0$ for the early parts of training, after which it stabilizes to near 0, indicating limited gradient interference between tasks. Even after employing Aligned-MTL \cite{senushkin2023independent}, as shown in \cref{fig:left_plot_gradient}, only smaller fluctuations in similarity values are observed at the beginning, while the final stabilization near zero remains unchanged, suggesting that our model suffers negligible gradient conflicts.

\subsection{Attention Visualization}
We demonstrate our model's capacity to capture multiple semantic cues from a single video by visualizing its temporal attention mechanism. Unlike spatial attention in standard Vision Transformers, our model processes temporal frames as tokens. Consequently, its attention mechanism reveals the specific temporal segments prioritized for different tasks.

\begin{figure}[t]
    \centering
    \includegraphics[width=0.8\linewidth]{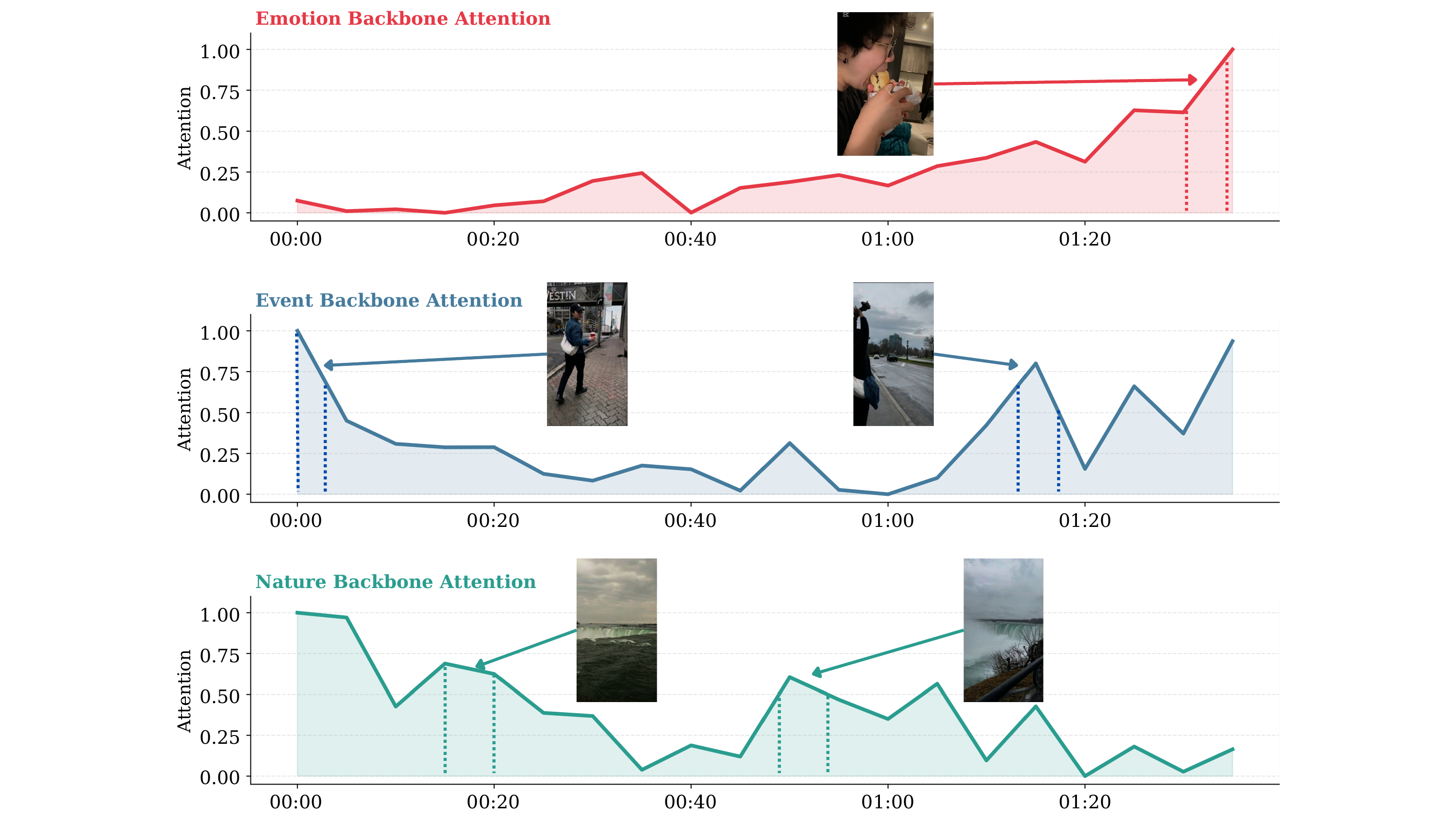}
    \caption{
Attention visualization of a personal vlog video.
The visualization shows temporal localization across different semantic perspectives.
The Emotion backbone highlights human-centric contentment, the Event backbone captures conversational interaction segments, and the Nature backbone focuses on environmental landmarks (Niagara Falls), illustrating the specialization of perspective-specific backbones.
}
    \label{fig:attn_viz}
\end{figure}

As Figure \ref{fig:attn_viz} shows, clear disentanglement between the three heads is observed, suggesting the model learns distinct, task-specific representations from the shared video backbone. We validate this semantic grounding by examining frames where attention values peaked. The model effectively localizes facial expressions for emotion, physical transitions for events, and landscape vistas for nature. Notably, the low cross-correlation between these head-specific distributions indicates that the model appears to partition the latent space to minimize task interference. Furthermore, the mechanism acts as a temporal filter, assigning low attention weights to uninformative segments, such as motion blur or camera resets, to focus purely on causally relevant evidence.

\section{Conclusion}
\label{sec:conclusion}

We revisit video highlight detection under realistic personal-style settings and identify the limitations of event-centric formulations. 
To address this gap, we introduce TRINITY, a multi-perspective benchmark that decomposes highlight saliency into event, emotion, and nature dimensions. 
We further propose a shared-backbone multi-expert architecture to model perspective-specific saliency.
Extensive experiments show strong performance on TRINITY and consistent gains on existing benchmarks, demonstrating improved generalization across heterogeneous highlight definitions.
We hope this work encourages more realistic and multi-dimensional formulations of highlight detection.

\appendix
\section{Details of Collection and Annotation of TRINITY}
\label{sec:appendix_data_details}

\subsection{Source Video Pool and Data Collection}
\begin{table}[H]
\centering
\caption{
Public datasets forming the source video pool used for constructing TRINITY.
}
\label{tab:emotion_nature_source_datasets}
\small
\setlength{\tabcolsep}{6pt}
\renewcommand{\arraystretch}{1.1}
\begin{tabular}{l c}
\toprule
\textbf{Official Name} & \textbf{\#Videos} \\
\midrule
QuerYD\cite{oncescu2021queryd} & 2,858 \\
TACoS\cite{rohrbach2014coherent} & 273 \\
Mr.\ HiSum\cite{sul2023mr} & 27,845 \\
DiDeMo\cite{anne2017localizing} & 10,463 \\
InternVid-VTime\cite{huang2024vtimellm} & 45,911 \\
CosMo-Cap\cite{wang2024cosmo} & 68,057 \\
QVHighlights\cite{lei2021detecting} & 12,562 \\
Charades-STA\cite{gao2017tall} & 9,848 \\
ActivityNet Captions\cite{krishna2017dense} & 13,545 \\
HiREST\cite{zala2023hierarchical} & 4,708 \\
\midrule
\textbf{Total} & \textbf{196,070} \\
\bottomrule
\end{tabular}
\end{table}

The three highlight perspectives in TRINITY are constructed using different data collection strategies due to the varying difficulty of annotation. 
Event-driven highlights are particularly challenging to annotate reliably at scale, as they require identifying semantically meaningful events with strong collective attention signals. 
To ensure statistical reliability, the Event perspective in TRINITY is derived entirely from Mr.~HiSum\cite{sul2023mr}, where replay statistics provide a large-scale and objective proxy for event-driven highlights.

In contrast, Emotion-driven and Nature-driven highlights can be annotated more directly based on visual cues such as facial expressions or scenic aesthetics. 
Therefore, these two perspectives are constructed from a large pool of publicly available video datasets. 
As summarized in Table~\ref{tab:emotion_nature_source_datasets}, the source pool consists of ten widely used datasets, including QuerYD\cite{oncescu2021queryd}, TACoS\cite{rohrbach2014coherent}, Mr.~HiSum\cite{sul2023mr}, DiDeMo\cite{anne2017localizing}, InternVid-VTime\cite{huang2024vtimellm}, CosMo-Cap\cite{wang2024cosmo}, QVHighlights\cite{lei2021detecting}, Charades-STA\cite{gao2017tall}, ActivityNet Captions\cite{krishna2017dense}, and HiREST\cite{zala2023hierarchical}. 
In total, the collection contains 196,070 videos spanning diverse domains such as daily activities, web videos, instructional content, and query-based video retrieval benchmarks. 
Emotion-driven and scenery-driven highlight peaks are annotated from videos sampled from this unified pool, ensuring diverse visual contexts while reducing dataset-specific biases.

\subsection{Cross-Perspective Overlap and Orthogonality Analysis}
\label{subsec:appendix_orthogonality}

Since the event-centric annotations in TRINITY originate from the Mr.~HiSum \cite{sul2023mr} source pool, we conduct a fine-grained structural and temporal overlap analysis specifically on this shared video collection. This ensures a consistent, co-annotated video pool to precisely measure cross-perspective complementarity and view sparsity.

Table~\ref{tab:orthogonality_combined} presents the structural overlap statistics across the perspectives within this subset. Event-driven and emotion-driven highlights show limited alignment with a temporal IoU of 0.034 across 3,096 videos, where emotional peaks are significantly shorter and exhibit much lower temporal coverage (4.35\,s, 5.0\%) compared to event segments (10.84\,s, 15.5\%). The overlap between event-driven and scenery-driven highlights is slightly higher (IoU=0.052 over 2,814 videos), capturing instances where striking landscapes coincide with narrative milestones. 

Crucially, emotion-driven and scenery-driven highlights rarely co-occur, yielding a negligible temporal IoU of 0.002, with only 78 videos containing valid annotations for both. These microscopic results demonstrate that the three perspectives capture complementary highlight signals with substantially independent temporal structures, confirming that TRINITY introduces largely orthogonal saliency cues that cannot be covered by any single perspective alone.

\begin{table}[H]
  \centering
  \caption{
Statistical analysis of highlight structures and temporal overlap across three perspectives on the Mr.~HiSum dataset pool.
\textit{Pair} denotes the compared perspectives (Ev: Event, Em: Emotion, Na: Nature).
\textit{Overlap} reports the temporal IoU between highlight segments.
$N_{\text{vids}}$ denotes the number of videos containing both annotations.
\textit{Peaks}, \textit{Dur.}, and \textit{Cov.} denote the average number of peaks, duration (seconds), and temporal coverage per video, respectively.
Values are reported as A/B format.
}
  \label{tab:orthogonality_combined}
  \footnotesize
  \setlength{\tabcolsep}{4pt}
  \renewcommand{\arraystretch}{1.15}
  \begin{tabular}{l c c c c c c}
    \toprule
    Dataset & Pair & Overlap & $N_{\text{vids}}$ & Peaks (\#) & Dur. (s) & Cov. (\%) \\
    \midrule
    \multirow{4}{*}{Mr.~HiSum}
    & Ev--Em & 0.034 & 3096 & 3.0/2.3 & 10.84/4.35 & 15.5/5.0 \\
    & Ev--Na & 0.052 & 2814 & 3.1/3.0 & 9.65/8.65 & 15.5/14.1 \\
    & Em--Na & 0.002 & 78 & 1.5/2.3 & 3.8/7.5 & 3.0/8.9 \\
    & Ev--Em--Na & N/A & 78 & 2.9/1.5/2.3 & 10.52/3.76/7.47 & 15.5/3.0/8.9 \\
    \bottomrule
  \end{tabular}
\end{table}

\subsection{Emotion Annotation Statistics}

\begin{table}[H]
\centering
\caption{Emotion distribution of extracted peaks (by peak count).}
\label{tab:emotion_dist}
\setlength{\tabcolsep}{8pt}
\renewcommand{\arraystretch}{1.15}
\begin{tabular}{l r r}
\toprule
\textbf{Emotion} & \textbf{\#Peaks} & \textbf{Ratio (\%)} \\
\midrule
Happiness & 27{,}737 & 86.55 \\
Sadness   & 3{,}408  & 10.63 \\
Surprise  & 452      & 1.41 \\
Anger     & 356      & 1.11 \\
Disgust   & 67       & 0.21 \\
Fear      & 26       & 0.08 \\
\midrule
Total     & 32{,}046 & 100.00 \\
\bottomrule
\end{tabular}
\end{table}

Table~\ref{tab:emotion_dist} summarizes the distribution of emotion categories among the extracted emotion-driven highlight peaks. 
The dataset is dominated by \textit{Happiness}, which accounts for 86.55\% of all annotated peaks. 
This predominance reflects the fact that positive emotional expressions such as smiling and laughter are both more visually salient and more frequently captured in web videos.

Negative emotions such as \textit{Sadness}, \textit{Anger}, and \textit{Disgust} appear less frequently, while highly specific expressions such as \textit{Fear} and \textit{Surprise} occur only in a small number of segments. 
Overall, the distribution exhibits a clear long-tail pattern, where a few common emotions account for the majority of peaks while several rarer categories remain sparsely represented.

Such imbalance is consistent with real-world video content and highlights the challenge of learning robust emotion-driven highlight detection across diverse affective expressions.

\subsection{Checking of Annotations}
\paragraph{Effect of Landscape Threshold.}

We evaluate the impact of different landscape thresholds $\theta_v^{\text{land}}$ on scenic video classification accuracy. 
A manually verified subset consisting of 100 videos is used for evaluation, including 80 scenic videos and 20 non-scenic videos. 
The person threshold is fixed to $\theta_v^{\text{per}} = 0.3$, while $\theta_v^{\text{land}}$ varies from 0.1 to 0.6.

For $\theta_v^{\text{land}} \in \{0.1, 0.2, 0.3, 0.4, 0.5\}$, 2 non-scenic videos are incorrectly classified as scenic (false positives), and 1 scenic video is incorrectly classified as non-scenic (false negative). 
When $\theta_v^{\text{land}} = 0.6$, 2 non-scenic videos are misclassified as scenic, while 16 scenic videos are misclassified as non-scenic.

\begin{table}[H]
  \centering
  \caption{Influence of landscape threshold $\theta_v^{\text{land}}$ on scenic video classification ($\theta_v^{\text{per}}$ is fixed to 0.3). 
  The evaluation subset contains 80 scenic and 20 non-scenic videos. 
  Accuracy is computed over all 100 videos.}
  \label{tab:landscape_threshold}
  \renewcommand{\arraystretch}{1.2}
  \setlength{\tabcolsep}{6pt}
  \begin{tabular}{c|c|c|c}
      \toprule
      \textbf{$\theta_v^{\text{land}}$} 
      & \textbf{False Positives} 
      & \textbf{False Negatives} 
      & \textbf{Accuracy} \\
      \midrule

      0.1 & 2 & 1 & 97\% \\
      0.2 & 2 & 1 & 97\% \\
      0.3 & 2 & 1 & 97\% \\

      \midrule
      \rowcolor[HTML]{F2F2F2}
      \textbf{0.4} & \textbf{2} & \textbf{1} & \textbf{97\%} \\
      \midrule

      0.5 & 2 & 1 & 97\% \\
      0.6 & 2 & 16 & 82\% \\

      \bottomrule
  \end{tabular}
\end{table}

Accuracy is computed as the ratio of correctly classified videos over the total 100 videos. 

These results indicate that the landscape threshold controls the minimum required proportion of scenic frames within a video. 
When $\theta_v^{\text{land}}$ is too small, videos with only a small fraction of scenic content may still be classified as scenic, weakening the purity of the dataset. 
When $\theta_v^{\text{land}}$ is too large, the criterion becomes overly strict, causing many genuinely scenic videos to be rejected. 

Considering both classification accuracy and the balance between inclusiveness and strictness, we select $\theta_v^{\text{land}} = 0.4$ as the default setting in TRINITY-Nature.

\paragraph{Annotation Consistency Analysis On Emotion Labels.}
To validate the reliability of the emotion-driven highlight annotations, we perform a human audit on 50 randomly sampled segments from the TRINITY-Emotion dataset with three independent annotators. Each annotator provides a binary judgment on whether each emotion label is correct: $y_t^i = 1$ if annotator $i$ considers the emotion label correct, else $y_t^i = 0$, $i = 1,2,3$.

For each segment, majority voting is applied across the three annotators to obtain the human consensus judgment: $y_\text{majority} = 1$ if $\sum_{i=1}^3 y_t^i \ge 2$, else $y_\text{majority} = 0$.

The human-validated accuracy of the emotion labels is computed as $\text{Accuracy} = \frac{\sum_{t=1}^{50} \mathbf{1}[y_\text{majority} = 1]}{50} \times 100\%$, where 47 out of 50 segments satisfy $y_\text{majority}=1$, yielding an accuracy of 94\%.

To measure agreement among annotators, we define the inter-annotator agreement rate as the proportion of segments where all three annotators unanimously agree (all 1 or all 0): $\text{Inter-Annotator Agreement} = \frac{\#\text{segments with } y_1 = y_2 = y_3}{50} \times 100\%$, which results in 46/50 = 92\%.

The results indicate high labeling consistency across annotators and reliability of the emotion-driven highlight annotations.

\paragraph{Annotation Consistency Analysis On Scenic Validity.} 
To evaluate the accuracy of scenic segment localization, we randomly sample 50 videos from TRINITY-Nature and let three independent annotators label scenic frames at 1\,fps. 
For each video, a human consensus set of scenic segments is obtained via majority voting, \ie, a frame is considered scenic if at least two annotators mark it as such. 
For each video, we then compute the Intersection-over-Union (IoU) between the human consensus segments and the labeled ground-truth scenic segments:
\begin{equation}
\text{IoU}_{\text{label-human}} = \frac{|S_{\text{consensus}} \cap S_{\text{GT}}|}{|S_{\text{consensus}} \cup S_{\text{GT}}|},
\end{equation}
where $S_{\text{consensus}}$ is the set of human consensus scenic frames and $S_{\text{GT}}$ is the set of ground-truth scenic frames. 
We also compute the pairwise inter-annotator IoU for each video:
\begin{equation}
\text{IoU}_{\text{A}_i\text{A}_j} = \frac{|S_{\text{A}_i} \cap S_{\text{A}_j}|}{|S_{\text{A}_i} \cup S_{\text{A}_j}|}, \quad i,j \in \{1,2,3\},~i<j,
\end{equation}
where $S_{\text{A}_i}$ denotes the scenic frames labeled by annotator $i$. 
Finally, we report the average IoU across all videos. 
The average label-human IoU is 0.89727, while the average inter-annotator IoU is 0.89744, indicating that the labeled scenic localization aligns with human judgment almost as well as different annotators agree among themselves.

\paragraph{Annotation Consistency Analysis On Aesthetic Ordering.} 
To assess the accuracy of model-selected scenic highlights, we use the same 50 videos sampled for the scenic validity evaluation. 
For each video, each annotator examines labeled highlight segments and makes a binary decision: 1 if the labeled highlight segments are correct, 0 otherwise. 
A majority vote over the three annotators determines whether the highlight segments of a video are correctly labeled. 
The Accuracy of the label is then computed as the fraction of videos where highlights are confirmed correct by majority vote:
\begin{equation}
\text{Accuracy} = \frac{N_{\text{correct}}}{N_{\text{total}}} = \frac{47}{50} = 94\%.
\end{equation}
We also measure inter-annotator agreement for highlight validation by counting the number of videos where all three annotators give the same judgment (either all 0 or all 1):
\begin{equation}
\text{Inter-annotator agreement} = \frac{N_{\text{same}}}{N_{\text{total}}} = \frac{45}{50} = 90\%.
\end{equation}
These results suggest that the model's selection of aesthetic highlights is largely consistent with human judgment and that annotators themselves achieve high agreement.

\section{Additional Ablation Studies and Further Analysis}

\subsection{Sensitivity of Loss Weights}
Our objective function is defined as the unweighted sum of the three task-specific losses. While sophisticated dynamic weighting strategies have been proposed to balance multi-task learning \cite{Chen2018GradNorm,Guo2018DynamicTP,Sener2018MultiTaskLA}, we found that a uniform weighting scheme provides the most stable convergence and superior performance for our specific architecture and negligible differences with manual tuning. We evaluate several alternative weighting approaches for comparison, the results of which are summarized in Table \ref{tab:weightlossstrat}. 

\begin{table}[H]
  \centering
  \caption{Comparison of various weight loss strategies.}
  \label{tab:weightlossstrat}
  \renewcommand{\arraystretch}{1.2}
  \setlength{\tabcolsep}{4pt} 
  \resizebox{0.9\textwidth}{!}{
    \begin{tabular}{l|cc|cc|cc|cc}
      \toprule
      \multirow{3}{*}{Method} 
      & \multicolumn{2}{c|}{TRINITY-Nature} 
      & \multicolumn{2}{c|}{TRINITY-Emotion} 
      & \multicolumn{2}{c|}{TRINITY-Event} 
      & \multicolumn{2}{c}{Average} \\
      \cmidrule(lr){2-3} \cmidrule(lr){4-5} \cmidrule(lr){6-7} \cmidrule(lr){8-9}
      & $\mathrm{mAP}_{15}$ 
      & $\mathrm{mAP}_{50}$ 
      & $\mathrm{mAP}_{15}$ 
      & $\mathrm{mAP}_{50}$ 
      & $\mathrm{mAP}_{15}$ 
      & $\mathrm{mAP}_{50}$ 
      & $\mathrm{mAP}_{15}$ 
      & $\mathrm{mAP}_{50}$ \\
      \midrule

      Uncertainty \cite{Kendalletal2018}
      & 50.48 & 68.45
      & 74.67 & 61.75
      & 38.15 & 68.56 
      & 54.43 & 66.25 \\

      Large event weight
      & 58.43 & 67.65
      & 67.26 & 71.77
      & 40.46 & 69.55
      & 55.38 & 69.66 \\

      Small event weight
      & 58.91 & 69.37 
      & 66.46 & 72.08
      & 40.57 & 69.56
      & 55.31 & 70.34 \\

      \midrule
      \rowcolor[HTML]{F2F2F2}
      Uniform 
      & 58.74 & 68.05 
      & 67.42 & 71.32 
      & 40.98 & 69.06 
      & 55.71 & 69.48 \\
      \bottomrule
    \end{tabular}
  }
\end{table}

Dynamic methods such as the homoscedastic uncertainty method \cite{Kendalletal2018}, introduces meaningful variation in our loss weights, but fails to perform as well as the uniform weights.

Observing that our model performs the worst on the event data, we ran experiments with different event weight values ($\lambda=2$ for large event weight, $\lambda=0.5$ for small event weight). Despite these changes, there are no significant changes in the result.  

This suggests that tedious hyperparameter tuning for our method is unnecessary and yields no significant improvement over our simple, uniform weighting method. This alleviates the burden of having to find the most optimal loss weights via tedious tuning.

\subsection{Training methods}
\begin{table}[htbp]
  \centering
  \caption{Comparison of various weight loss strategies with average performance across datasets.}
  \label{tab:trainingmethods}
  \renewcommand{\arraystretch}{1.2}
  \setlength{\tabcolsep}{4.5pt} 
  \resizebox{0.9\textwidth}{!}{
    \begin{tabular}{l|cc|cc|cc|cc}
      \toprule
      \multirow{3}{*}{\textbf{Method}} 
      & \multicolumn{2}{c|}{\textbf{TRINITY-Nature}} 
      & \multicolumn{2}{c|}{\textbf{TRINITY-Emotion}} 
      & \multicolumn{2}{c|}{\textbf{TRINITY-Event}} 
      & \multicolumn{2}{c}{\textbf{Average}} \\
      \cmidrule(lr){2-3} \cmidrule(lr){4-5} \cmidrule(lr){6-7} \cmidrule(lr){8-9}
      & $\mathrm{mAP}_{15}$ 
      & $\mathrm{mAP}_{50}$ 
      & $\mathrm{mAP}_{15}$ 
      & $\mathrm{mAP}_{50}$ 
      & $\mathrm{mAP}_{15}$ 
      & $\mathrm{mAP}_{50}$ 
      & $\mathbf{mAP}_{15}$ 
      & $\mathbf{mAP}_{50}$ \\
      \midrule

      Single Backbone (Seq.)
      & 46.47 & 64.76
      & 66.34 & 72.35
      & 33.86 & 64.66 
      & 48.89 & 67.26 \\

      Single Backbone (R.R.)
      & 44.41 & 64.35
      & 65.20 & 75.71 
      & 37.97 & 66.86 
      & 49.19 & 68.97 \\

      Ours (Sequential) 
      & 58.71 & 69.20
      & 68.30 & 71.15
      & 39.67 & 68.97 
      & 55.56 & \textbf{69.77} \\

      \midrule
      \rowcolor[HTML]{F2F2F2}
      \textbf{Ours (Original)} 
      & 58.74 & 68.05 
      & 67.42 & 71.32 
      & 40.98 & 69.06 
      & \textbf{55.71} & 69.48 \\
      \bottomrule
    \end{tabular}
  }
\end{table}

To mitigate catastrophic forgetting across the diverse domains within our Trinity dataset, we implement an interleaved training strategy. We process mini-batches from the three constituent tasks in a round-robin fashion, accumulating gradients across all three before performing a single backpropagation step. This synchronized update ensures that the model's parameters are optimized against a balanced objective, preventing the weights from overfitting to any single task's distribution in isolation.

We compare our proposed method with two alternatives. The first is a sequential training method for a single backbone architecture, which uses one single transformer based backbone with three different heads for each of the three tasks. The second is another sequential training method with our own proposed model architecture. These two alternatives are compared with our proposed round-robin training method with our model architecture. 

As seen in Table \ref{tab:trainingmethods}, the single backbone trained sequentially performs very poorly compared to our proposed method. With all the data passing through the common backbone in a sequential fashion, one task type at a time, the backbone forgets what it has learned from previous tasks, showing signs of catastrophic forgetting. When run with round robin fashion however, the results are noticeably better (S4).  

However, our method shows resilience against catastrophic forgetting and only shows minimal degradation in performance in the more punishing $\mathrm{mAP}_{\rho=15\%}$ metric when using sequential training. This can be attributed to the presence of private backbones which are fitted to specific tasks. Even though the shared backbone may be exposed to sequential training, the private backbones are able to retain task specific information, and together with the global information learned by the shared backbone, output great results. As such, our architecture combined with the round robin data loading style, which allows for even training across the different tasks, produces the best $\mathrm{mAP}_{\rho=15\%}$ result. 

\subsection{Analysis on View Co-occurrence and MixBatch Strategy}
\label{subsec:appendix_cooccurrence_mixbatch}

TRINITY is designed around a view-sparsity framework, meaning that videos do not necessarily exhibit all three highlight perspectives (Event, Emotion, and Nature) simultaneously. Instead, saliency within each perspective can manifest independently or partially across the dataset. To thoroughly evaluate our model's capability under a dense-view scenario where all perspectives overlap, we analyze a strict co-occurrence subset comprising 78 videos that concurrently contain valid annotations for all three dimensions. Furthermore, we investigate an alternative multi-task training variant, \textit{MixBatch}, which enforces intra-batch perspective mixing by blending samples from different tasks within a single mini-batch, rather than isolating tasks per batch as in our default round-robin setup.

The ablation results are summarized in Table~\ref{tab:appendix_cooccurrence_mixbatch}. When evaluated on the highly constrained 78-video co-occurrence subset, our model consistently maintains robust performance across all perspectives, demonstrating its proficiency in executing distinct, perspective-specific predictions even under dense, multi-cue contexts. 

Regarding the batching configurations, the \textit{MixBatch} strategy yields marginal performance gains on the TRINITY-Emotion and TRINITY-Event benchmarks. However, it incurs a significant performance degradation on the TRINITY-Nature dimension, with the $\mathrm{mAP}_{15}$ dropping sharply from 58.74\% to 42.19\%. This implies that while intra-batch perspective mixing is technically feasible and acceptable for action or affect-driven tasks, it tends to disrupt the fine-grained visual representations required for scenic aesthetic assessment. Consequently, our default perspective-isolated batching formulation remains the most optimal and balanced choice for stabilizing multi-perspective highlight learning.

\begin{table}[H]
  \centering
  \caption{Co-occurrence and training-strategy ablations on TRINITY.}
  \label{tab:appendix_cooccurrence_mixbatch}
  \small
  \setlength{\tabcolsep}{5pt}
  \renewcommand{\arraystretch}{1.15}
  \resizebox{0.95\textwidth}{!}{
    \begin{tabular}{llcccccc}
      \toprule
      \multirow{2}{*}{\textbf{Setting}} 
      & \multirow{2}{*}{\textbf{Eval. Set}}
      & \multicolumn{2}{c}{\textbf{TRINITY-Nature}} 
      & \multicolumn{2}{c}{\textbf{TRINITY-Emotion}} 
      & \multicolumn{2}{c}{\textbf{TRINITY-Event}} \\
      \cmidrule(lr){3-4} \cmidrule(lr){5-6} \cmidrule(lr){7-8}
      & & $\mathrm{mAP}_{15}$ & $\mathrm{mAP}_{50}$
      & $\mathrm{mAP}_{15}$ & $\mathrm{mAP}_{50}$
      & $\mathrm{mAP}_{15}$ & $\mathrm{mAP}_{\rho=50\%}$ \\
      \midrule
      Ours
      & Full test
      & 58.74 & 68.05
      & 67.42 & 71.32
      & 40.98 & 69.06 \\

      Ours
      & 78-video subset
      & 46.66 & 66.36
      & 47.50 & 62.87
      & 36.70 & 67.89 \\

      Ours w/ MixBatch
      & Full test
      & 42.19 & 63.54
      & 63.61 & 75.40
      & 39.06 & 67.60 \\
      \bottomrule
    \end{tabular}
  }
\end{table}

\subsection{Hyperparameters}
We list down the hyperparameters used for finetuning on these respective datasets. For all finetuning, the same early stopping as mentioned in the main paper was utilized. 

\begin{table}[htbp]
\centering
\caption{Hyperparameter settings for VEATIC and YouTubeHighlights datasets.}
\label{tab:hyperparameters}
\begin{tabular}{lcc}
\hline
\textbf{Hyperparameters} & \textbf{VEATIC} & \textbf{YouTubeHighlights} \\ \hline
learning rate        & 3e-4         & 1e-4                    \\
l2 regularization                 & 3.9         & 1.8                    \\
batch size             & 4         & 16                    \\
dropout ratio          & 0.43         & 0.42                    \\
epochs                  & 100         & 100                    \\ \hline
\end{tabular}
\end{table}

\section{Case Visualization}
\label{sec:appendix_vis}

We present four representative cases to qualitatively analyze the behavior of our model under different video scenarios. 
The four examples correspond to: (1) a skateboarding practice video dominated by human actions, 
(2) a crafting vlog that records the process of making decorative leaf flowers, 
(3) an indoor dance competition performance, and 
(4) a tool-making tutorial consisting mainly of close-up manual operations. 

For each case, we visualize the predicted highlight scores from the three TRINITY perspectives: Event, Emotion, and Nature. 
These examples illustrate how the perspective-specific experts respond selectively according to the semantic content of the video.

\clearpage
\begin{figure*}[p]
    \centering
    \begin{subfigure}[t]{0.95\textwidth}
        \centering
        \includegraphics[width=\linewidth]{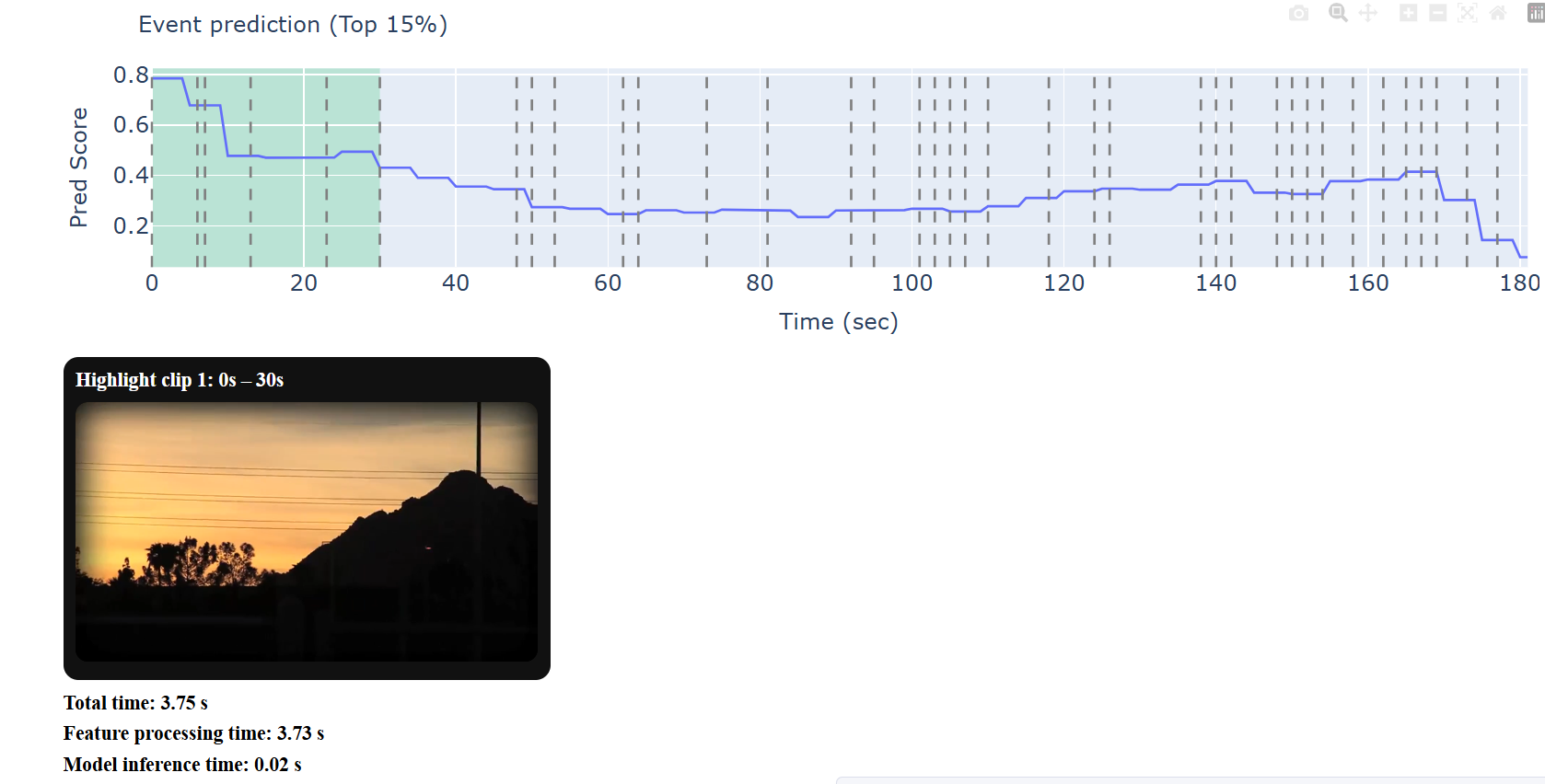}
        \caption{Event perspective prediction.}
        \label{fig:case1_event}
    \end{subfigure}
    
    \begin{subfigure}[t]{0.95\textwidth}
        \centering
        \includegraphics[width=\linewidth]{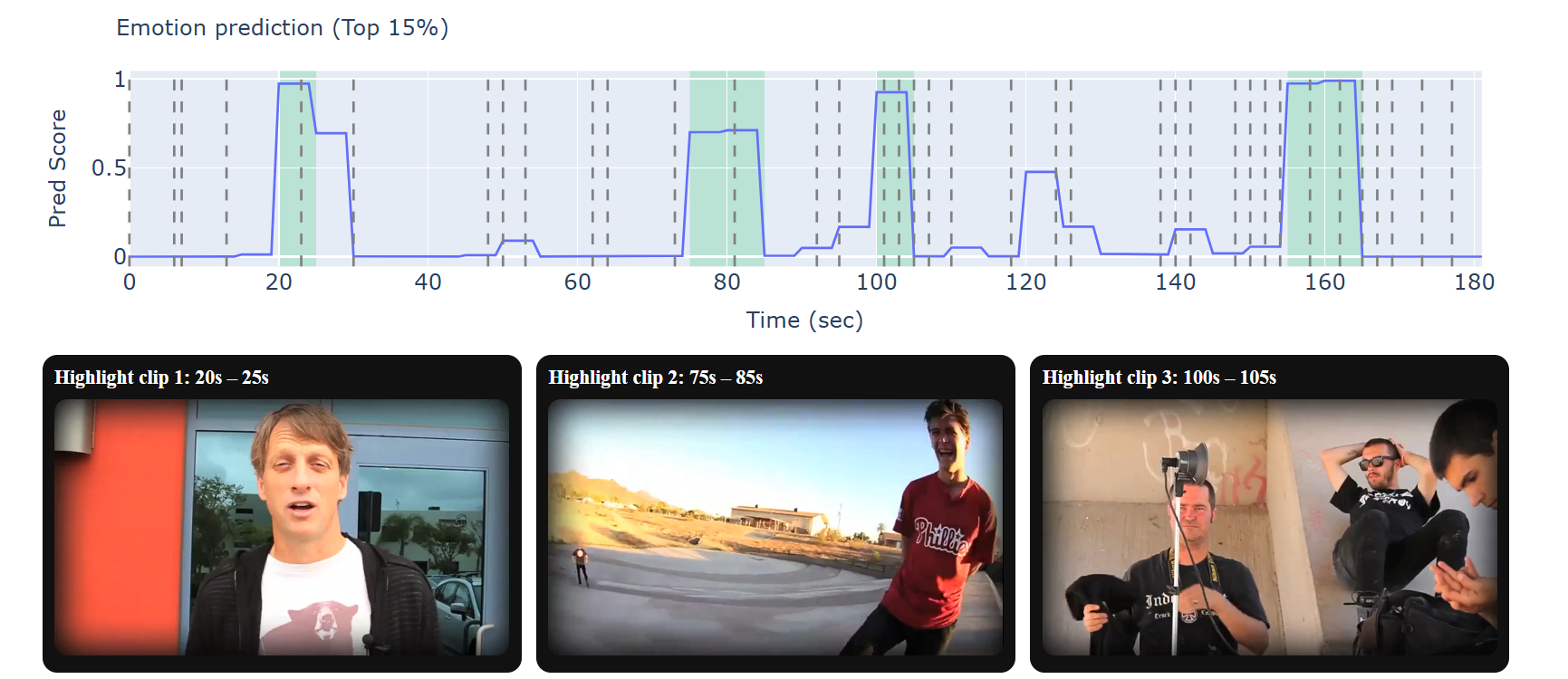}
        \caption{Emotion perspective prediction.}
        \label{fig:case1_emotion}
    \end{subfigure}
    
    \begin{subfigure}[t]{0.95\textwidth}
        \centering
        \includegraphics[width=\linewidth]{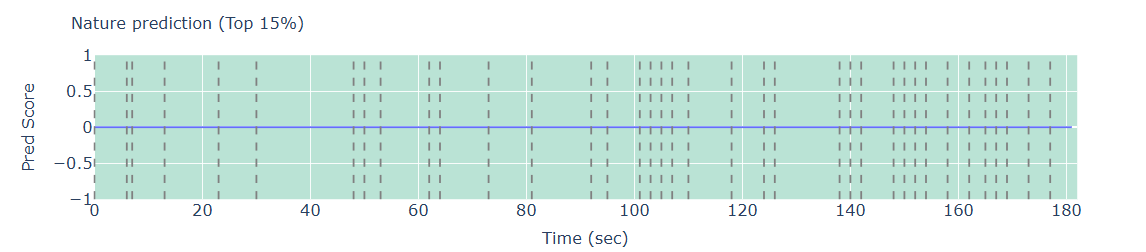}
        \caption{Nature perspective prediction.}
        \label{fig:case1_nature}
    \end{subfigure}

    \caption{
    \textbf{Case 1: Skateboarding practice video.}
    The video is dominated by human actions rather than scenic content. 
    The Event branch captures the temporal structure of skateboarding attempts, while the Emotion branch produces peaks around visible facial reactions. 
    The Nature branch remains largely inactive due to the absence of natural scenery.
    Predicted highlight scores are shown as curves, and shaded regions indicate ground-truth highlights.
    }
    \label{fig:case1_vis}
\end{figure*}

\clearpage
\begin{figure*}[p]
    \centering
    \begin{subfigure}[t]{0.95\textwidth}
        \centering
        \includegraphics[width=\linewidth]{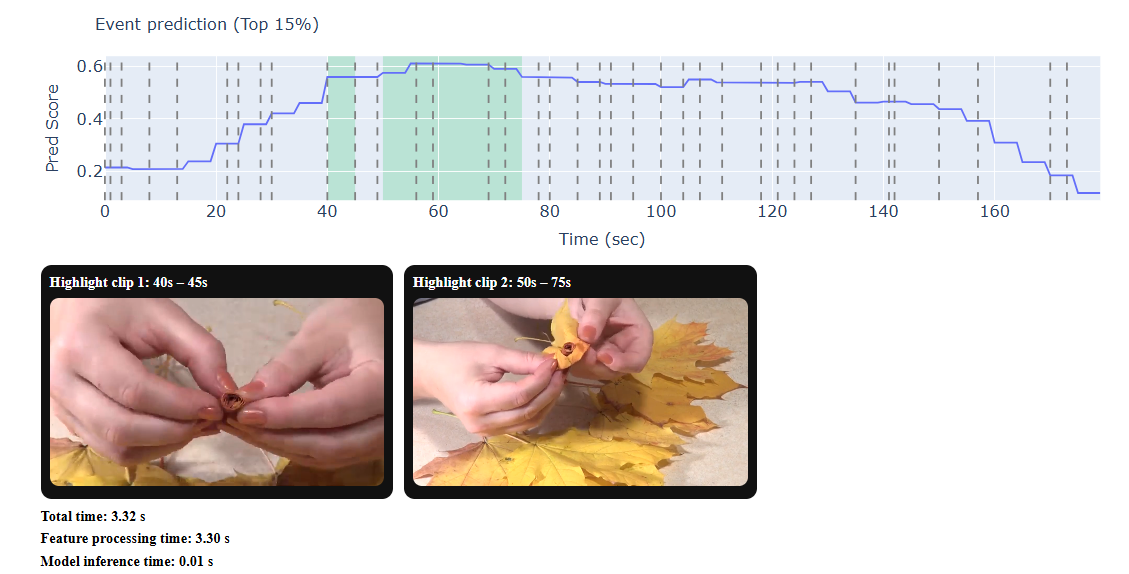}
        \caption{Event perspective prediction.}
        \label{fig:case2_event}
    \end{subfigure}
    
    \begin{subfigure}[t]{0.95\textwidth}
        \centering
        \includegraphics[width=\linewidth]{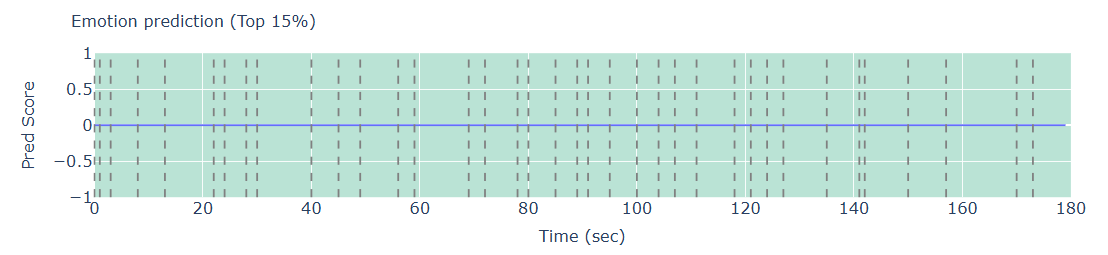}
        \caption{Emotion perspective prediction.}
        \label{fig:case2_emotion}
    \end{subfigure}
    
    \begin{subfigure}[t]{0.95\textwidth}
        \centering
        \includegraphics[width=\linewidth]{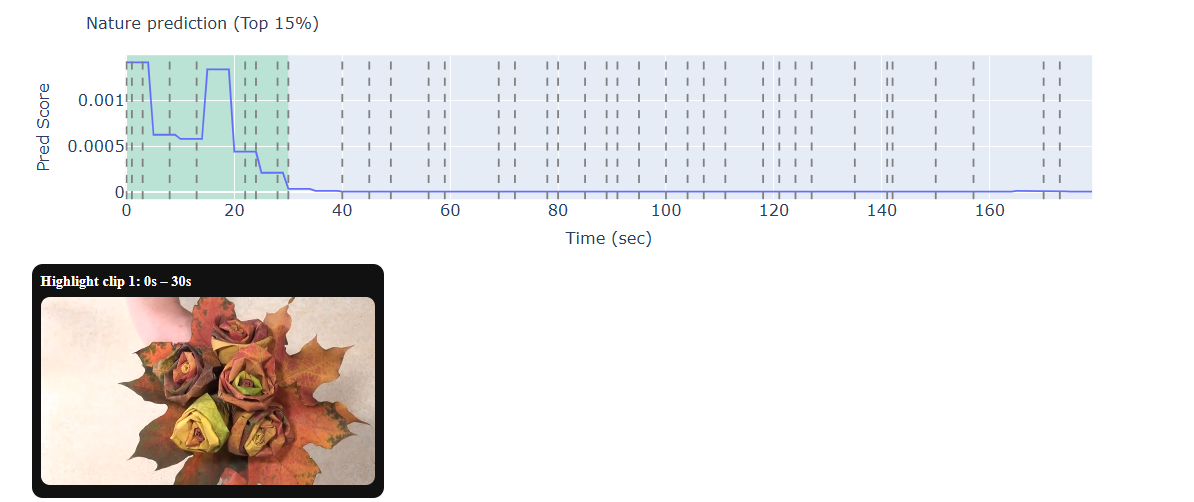}
        \caption{Nature perspective prediction.}
        \label{fig:case2_nature}
    \end{subfigure}

    \caption{
    \textbf{Case 2: Leaf-flower crafting vlog.}
    The video records the process of making decorative leaf flowers. 
    Since no faces appear in the video, the Emotion branch remains inactive. 
    The Nature branch highlights the aesthetically pleasing opening segment showing the completed leaf flower, while the Event branch captures the key crafting actions.
    Predicted highlight scores are shown as curves, and shaded regions indicate ground-truth highlights.
    }
    \label{fig:case2_vis}
\end{figure*}

\clearpage
\begin{figure*}[p]
    \centering
    \begin{subfigure}[t]{0.95\textwidth}
        \centering
        \includegraphics[width=\linewidth]{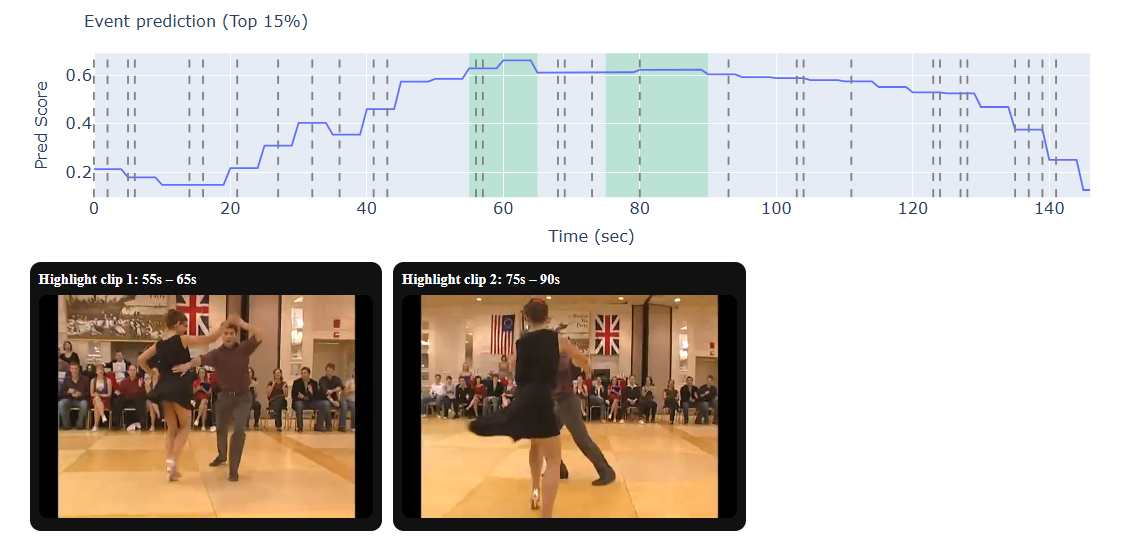}
        \caption{Event perspective prediction.}
        \label{fig:case3_event}
    \end{subfigure}
    
    \begin{subfigure}[t]{0.95\textwidth}
        \centering
        \includegraphics[width=\linewidth]{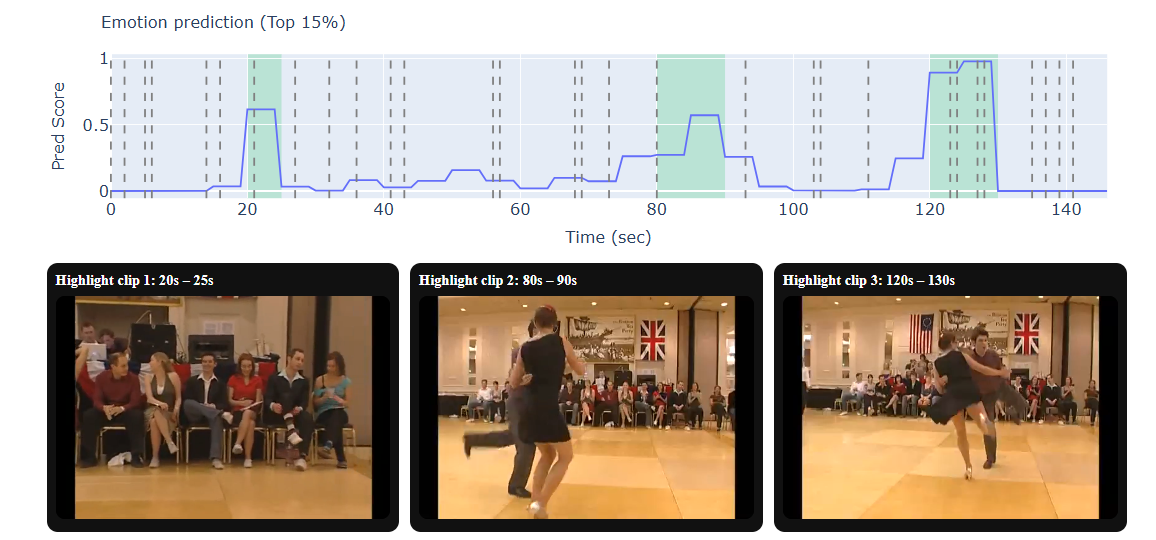}
        \caption{Emotion perspective prediction.}
        \label{fig:case3_emotion}
    \end{subfigure}
    
    \begin{subfigure}[t]{0.95\textwidth}
        \centering
        \includegraphics[width=\linewidth]{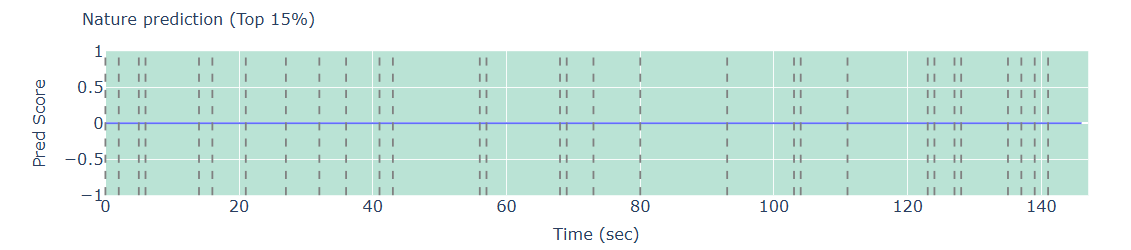}
        \caption{Nature perspective prediction.}
        \label{fig:case3_nature}
    \end{subfigure}

    \caption{
    \textbf{Case 3: Indoor dance competition performance.}
    The video takes place entirely indoors and therefore contains no natural scenery. 
    The Event branch captures key dance movements and performance highlights, while the Emotion branch responds to expressive moments and audience reactions. 
    The Nature branch remains inactive throughout the timeline.
    Predicted highlight scores are shown as curves, and shaded regions indicate ground-truth highlights.
    }
    \label{fig:case3_vis}
\end{figure*}

\clearpage
\begin{figure*}[p]
    \centering
    \begin{subfigure}[t]{0.95\textwidth}
        \centering
        \includegraphics[width=\linewidth]{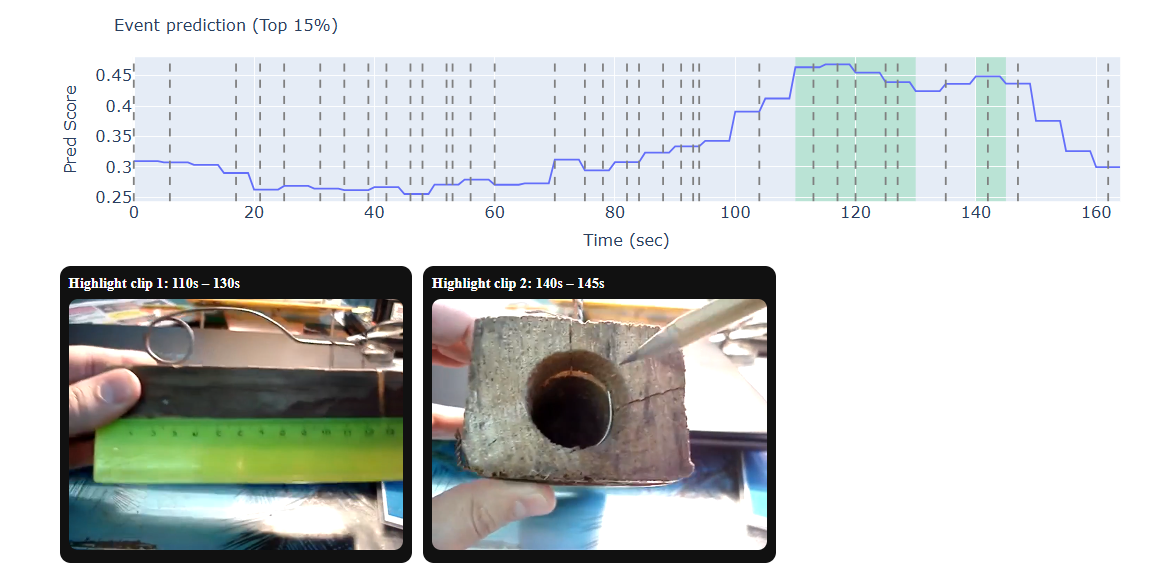}
        \caption{Event perspective prediction.}
        \label{fig:case4_event}
    \end{subfigure}
    
    \begin{subfigure}[t]{0.95\textwidth}
        \centering
        \includegraphics[width=\linewidth]{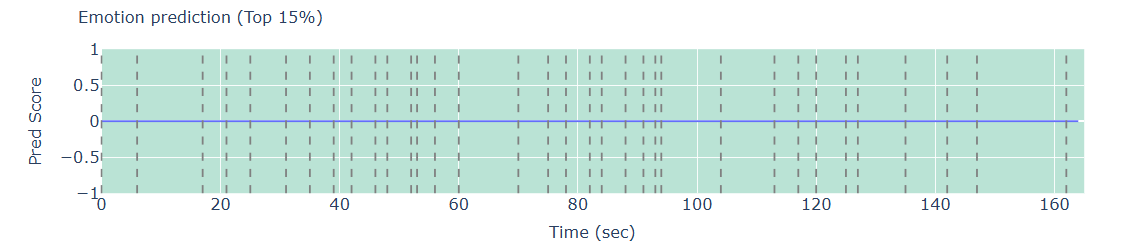}
        \caption{Emotion perspective prediction.}
        \label{fig:case4_emotion}
    \end{subfigure}
    
    \begin{subfigure}[t]{0.95\textwidth}
        \centering
        \includegraphics[width=\linewidth]{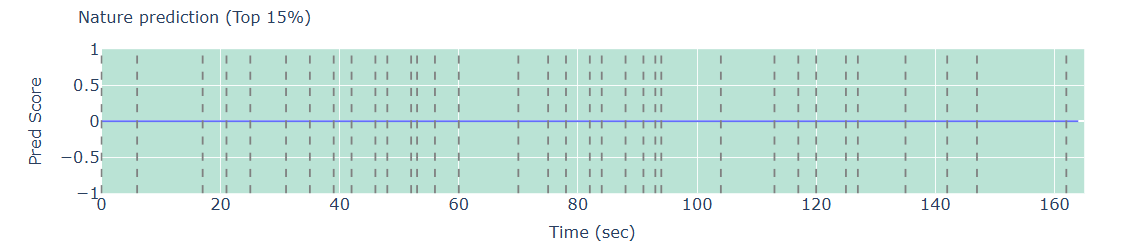}
        \caption{Nature perspective prediction.}
        \label{fig:case4_nature}
    \end{subfigure}

    \caption{
    \textbf{Case 4: Tool-making tutorial.}
    The video mainly consists of close-up manual operations during the construction of a tool. 
    Since there are no visible faces or natural scenery, both the Emotion and Nature branches remain inactive. 
    The Event branch instead captures the key procedural steps of the crafting process.
    Predicted highlight scores are shown as curves, and shaded regions indicate ground-truth highlights.
    }
    \label{fig:case4_vis}
\end{figure*}

\clearpage

\bibliographystyle{splncs04}
\bibliography{references,reference2}

\begin{thebibliography}{10}
\providecommand{\url}[1]{\texttt{#1}}
\providecommand{\urlprefix}{URL }
\providecommand{\doi}[1]{https://doi.org/#1}

\bibitem{alharbi2024effective}
Alharbi, F., Habib, S., Albattah, W., Jan, Z., Alanazi, M.D., Islam, M.:
  Effective video summarization using channel attention-assisted
  encoder--decoder framework. Symmetry  \textbf{16}(6), ~680 (2024)

\bibitem{badamdorj2021joint}
Badamdorj, T., Rochan, M., Wang, Y., Cheng, L.: Joint visual and audio learning
  for video highlight detection. In: ICCV. pp. 8127--8137 (2021)

\bibitem{Bai2025Qwen3VL}
Bai, S., Cai, Y., Chen, R., Chen, K., Chen, X., Cheng, Z., Deng, L., Ding, W.,
  Gao, C., Ge, C., Ge, W., Guo, Z., Huang, Q., Huang, J., Huang, F., Hui, B.,
  Jiang, S., Li, Z., Li, M., Li, M., Li, K., Lin, Z., Lin, J., Liu, X., Liu,
  J., Liu, C., Liu, Y., Liu, D., Liu, S., Lu, D., Luo, R., Lv, C., Men, R.,
  Meng, L., Ren, X., Ren, X., Song, S., Sun, Y., Tang, J., Tu, J., Wan, J.,
  Wang, P., Wang, P., Wang, Q., Wang, Y., Xie, T., Xu, Y., Xu, H., Xu, J.,
  Yang, Z., Yang, M., Yang, J., Yang, A., Yu, B., Zhang, F., Zhang, H., Zhang,
  X., Zheng, B., Zhong, H., Zhou, J., Zhou, F., Zhou, J., Zhu, Y., Zhu, K.:
  {Qwen3-VL} technical report (2025)

\bibitem{beedu2025hiersum}
Beedu, A., Essa, I.: {HierSum}: A global and local attention mechanism for
  video summarization. arXiv preprint arXiv:2504.18689  (2025)

\bibitem{cha2021conflict}
Cha, C., You, T., Yeh, R.A., Vitelli, S., Wang, S.H.: Conflict-averse gradient
  descent for multi-task learning. In: Advances in Neural Information
  Processing Systems. vol.~34, pp. 18803--18816 (2021)

\bibitem{Chang2025aha}
Chang, A., De~Melo, C., Lukin, S.M.: Aha! - predicting what matters next:
  Online highlight detection without looking ahead. In: NeurIPS (2025)

\bibitem{Chen2018GradNorm}
Chen, Z., Badrinarayanan, V., Lee, C.Y., Rabinovich, A.: {GradNorm}: Gradient
  normalization for adaptive loss balancing in deep multitask networks. In:
  ICML. pp. 794--803 (2018)

\bibitem{gao2017tall}
Gao, J., Sun, C., Yang, Z., Nevatia, R.: Tall: Temporal activity localization
  via language query. In: Proceedings of the IEEE international conference on
  computer vision. pp. 5267--5275 (2017)

\bibitem{grauman2022ego4d}
Grauman, K., Westbury, A., Byrne, E., Chavis, Z., Furnari, A., Girdhar, R.,
  Hamburger, J., Jiang, H., Liu, M., Liu, X., et~al.: {Ego4D}: Around the world
  in 3,000 hours of egocentric video. In: CVPR. pp. 18995--19012 (2022)

\bibitem{Guo2018DynamicTP}
Guo, M., Haque, A., Huang, D.A., Yeung, S., Fei-Fei, L.: Dynamic task
  prioritization for multi-task learning. In: ECCV. pp. 270--287 (2018)

\bibitem{gygli2016video2gif}
Gygli, M., Song, Y., Cao, L.: {Video2GIF}: Automatic generation of animated
  {GIFs} from video. In: CVPR. pp. 1001--1009 (2016)

\bibitem{anne2017localizing}
Hendricks, L.A., Wang, O., Shechtman, E., Sivic, J., Darrell, T., Russell, B.:
  Localizing moments in video with natural language. In: ICCV. pp. 5803--5812
  (2017)

\bibitem{hong2024csta}
Hong, S.E., Choi, G.M., Kim, J.H.: Csta: Cnn-based spatiotemporal attention for
  video summarization. In: Proceedings of the IEEE/CVF Conference on Computer
  Vision and Pattern Recognition (CVPR). pp. 18859--18868 (2024)

\bibitem{huang2024vtimellm}
Huang, B., Wang, X., Chen, H., Song, Z., Zhu, W.: {VTimeLLM}: Empower {LLM} to
  grasp video moments. In: CVPR. pp. 14271--14280 (2024)

\bibitem{islam2025unsupervised}
Islam, Z., Paul, S., Rochan, M.: Unsupervised video highlight detection by
  learning from audio and visual recurrence. In: Proc. WACV. pp. 8702--8711
  (2025)

\bibitem{jang2023knowing}
Jang, J., Park, J., Kim, J., Kwon, H., Sohn, K.: Knowing where to focus:
  Event-aware transformer for video grounding. In: ICCV. pp. 13846--13856
  (2023)

\bibitem{yolov8_ultralytics}
Jocher, G., Chaurasia, A., Qiu, J.: Ultralytics yolov8 (2023)

\bibitem{kahneman1993more}
Kahneman, D., Fredrickson, B.L., Schreiber, C.A., Redelmeier, D.A.: When more
  pain is preferred to less: Adding a better end. Psychological Science
  \textbf{4}(6),  401--405 (1993)

\bibitem{Ke2021MUSIQ}
Ke, J., Wang, Q., Wang, Y., Milanfar, P., Yang, F.: {MUSIQ}: Multi-scale image
  quality transformer. In: ICCV (2021)

\bibitem{Kendalletal2018}
Kendall, A., Gal, Y., Cipolla, R.: Multi-task learning using uncertainty to
  weigh losses for scene geometry and semantics. In: CVPR. pp. 7482--7491
  (2021)

\bibitem{Kim2025summdiff}
Kim, K., Hahm, J., Kim, S., Sul, J., Kim, B., Lee, J.: {SummDiff}: Generative
  modeling of video summarization with diffusion. In: ICCV. pp. 15096--15106
  (2025)

\bibitem{kirk2007understanding}
Kirk, D., Sellen, A., Harper, R., Wood, K.: Understanding videowork. In: Proc.
  CHI. pp. 61--70 (2007)

\bibitem{koutras2021combining}
Koutras, P., Maragos, P.: Combining global and local attention with positional
  encoding for video summarization. In: 2021 IEEE International Symposium on
  Multimedia (ISM). pp. 41--48 (2021)

\bibitem{krishna2017dense}
Krishna, R., Hata, K., Ren, F., Fei-Fei, L., Niebles, J.C.: Dense-captioning
  events in videos. In: ICCV. pp. 706--715 (2017)

\bibitem{lee2018deep}
Lee, H.B., Yang, E., Hwang, S.J.: Deep asymmetric multi-task feature learning.
  In: Dy, J., Krause, A. (eds.) Proceedings of the 35th International
  Conference on Machine Learning. Proceedings of Machine Learning Research,
  vol.~80, pp. 2956--2964. PMLR (2018)

\bibitem{lei2021detecting}
Lei, J., Berg, T.L., Bansal, M.: Detecting moments and highlights in videos via
  natural language queries. NeurIPS  \textbf{34},  11846--11858 (2021)

\bibitem{li2024unsupervised}
Li, T., Sun, Z., Xiao, X.: Unsupervised modality-transferable video highlight
  detection with representation activation sequence learning. IEEE TIP
  \textbf{33},  1911--1922 (2024)

\bibitem{lin2023univtg}
Lin, K.Q., Zhang, P., Chen, J., Pramanick, S., Gao, D., Wang, A.J., Yan, R.,
  Shou, M.Z.: {UniVTG}: Towards unified video-language temporal grounding. In:
  ICCV. pp. 2794--2804 (2023)

\bibitem{liu2022umt}
Liu, Y., Li, S., Wu, Y., Chen, C.W., Shan, Y., Qie, X.: {UMT}: Unified
  multi-modal transformers for joint video moment retrieval and highlight
  detection. In: CVPR. pp. 3042--3051 (2022)

\bibitem{McCloskey1989}
McCloskey, M., Cohen, N.J.: Catastrophic interference in connectionist
  networks: The sequential learning problem. In: Psychology of Learning and
  Motivation, vol.~24, pp. 109--165. Academic Press (1989)

\bibitem{garcia2018phd}
Garcia~del Molino, A., Gygli, M.: {PHD-GIFs}: Personalized highlight detection
  for automatic {GIF} creation. In: ACM MM. pp. 600--608 (2018)

\bibitem{moon2023query}
Moon, W., Hyun, S., Park, S., Park, D., Heo, J.P.: Query-dependent video
  representation for moment retrieval and highlight detection. In: CVPR. pp.
  23023--23033 (2023)

\bibitem{oncescu2021queryd}
Oncescu, A.M., Henriques, J.F., Liu, Y., Zisserman, A., Albanie, S.: {QueryD}:
  A video dataset with high-quality text and audio narrations. In: ICASSP. pp.
  2265--2269 (2021)

\bibitem{radford2021learning}
Radford, A., Kim, J.W., Hallacy, C., Ramesh, A., Goh, G., Agarwal, S., Sastry,
  G., Askell, A., Mishkin, P., Clark, J., et~al.: Learning transferable visual
  models from natural language supervision. In: ICML. pp. 8748--8763 (2021)

\bibitem{ren2024veatic}
Ren, Z., Ortega, J., Wang, Y., Chen, Z., Guo, Y., Yu, S.X., Whitney, D.:
  {VEATIC}: Video-based emotion and affect tracking in context dataset. In:
  Proc. WACV. pp. 4467--4477 (2024)

\bibitem{ren2024region}
Ren, Z., Wang, Y., Ke, T.W., Guo, Y., Yu, S.X., Whitney, D.: Region-based
  emotion recognition via superpixel feature pooling. In: CVPRW (2024)

\bibitem{rohrbach2014coherent}
Rohrbach, A., Rohrbach, M., Qiu, W., Friedrich, A., Pinkal, M., Schiele, B.:
  Coherent multi-sentence video description with variable level of detail. pp.
  184--195 (2014)

\bibitem{saigo2025enhancing}
Saigo, S., Hayami, T., Watanabe, H.: Enhancing continuous emotion recognition
  via visually diverse frame selection. In: Proc. IEEE Global Conference on
  Consumer Electronics (GCCE). pp. 1275--1278 (2025)

\bibitem{savchenko2023facial}
Savchenko, A.: Facial expression recognition with adaptive frame rate based on
  multiple testing correction. In: ICML. pp. 30119--30129 (2023)

\bibitem{savchenko2022classifying}
Savchenko, A.V., Savchenko, L.V., Makarov, I.: Classifying emotions and
  engagement in online learning based on a single facial expression recognition
  neural network. IEEE Trans. Affect. Comput.  (2022)

\bibitem{sellen2007life}
Sellen, A.J., Fogg, A., Aitken, M., Hodges, S., Rother, C., Wood, K.: Do
  life-logging technologies support memory for the past? {An} experimental
  study using {SenseCam}. In: Proc. CHI. pp. 81--90 (2007)

\bibitem{Sener2018MultiTaskLA}
Sener, O., Koltun, V.: Multi-task learning as multi-objective optimization. In:
  NeurIPS. pp. 525--535 (2018)

\bibitem{senushkin2023independent}
Senushkin, D., Belikov, N., Galimaldinov, A., Konushin, A.: Independent
  component alignment for multi-task learning. In: Proceedings of the IEEE/CVF
  Conference on Computer Vision and Pattern Recognition (CVPR). pp.
  20083--20093 (2023)

\bibitem{song2015tvsum}
Song, Y., Vallmitjana, J., Stent, A., Jaimes, A.: {TvSum}: Summarizing web
  videos using titles. In: CVPR. pp. 5179--5187 (2015)

\bibitem{su2024roformer}
Su, J., Lu, Y., Pan, S., Wen, B., Liu, Y.: Roformer: Enhanced transformer with
  rotary position embedding. arXiv preprint arXiv:2104.09864  (2021)

\bibitem{sul2023mr}
Sul, J., Han, J., Lee, J.: {Mr. HiSum}: A large-scale dataset for video
  highlight detection and summarization. In: NeurIPS (2023)

\bibitem{sun2024tr}
Sun, H., Zhou, M., Chen, W., Xie, W.: {TR-DETR}: Task-reciprocal transformer
  for joint moment retrieval and highlight detection. In: AAAI. vol.~38, pp.
  4998--5007 (2024)

\bibitem{sun2014ranking}
Sun, M., Farhadi, A., Seitz, S.: Ranking domain-specific highlights by
  analyzing edited videos. In: ECCV. pp. 787--802 (2014)

\bibitem{wang2024cosmo}
Wang, A.J., Li, L., Lin, K.Q., Wang, J., Lin, K., Yang, Z., Wang, L., Shou,
  M.Z.: {Cosmo}: Contrastive streamlined multimodal model with interleaved
  pre-training. arXiv preprint arXiv:2401.00849  (2024)

\bibitem{wang2024qwen2}
Wang, P., Bai, S., Tan, S., Wang, S., Fan, Z., Bai, J., Chen, K., Liu, X.,
  Wang, J., Ge, W., et~al.: {Qwen2-VL}: Enhancing vision-language model's
  perception of the world at any resolution. arXiv preprint arXiv:2409.12191
  (2024)

\bibitem{wei2022learning}
Wei, F., Wang, B., Ge, T., Jiang, Y., Li, W., Duan, L.: Learning pixel-level
  distinctions for video highlight detection. In: CVPR. pp. 3073--3082 (2022)

\bibitem{xu2021cross}
Xu, M., Wang, H., Ni, B., Zhu, R., Sun, Z., Wang, C.: Cross-category video
  highlight detection via set-based learning. In: ICCV. pp. 7970--7979 (2021)

\bibitem{yao2016highlight}
Yao, T., Mei, T., Rui, Y.: Highlight detection with pairwise deep ranking for
  first-person video summarization. In: CVPR. pp. 982--990 (2016)

\bibitem{ye2021temporal}
Ye, Q., Shen, X., Gao, Y., Wang, Z., Bi, Q., Li, P., Yang, G.: Temporal cue
  guided video highlight detection with low-rank audio-visual fusion. In: ICCV.
  pp. 7950--7959 (2021)

\bibitem{yu2020gradient}
Yu, T., Kumar, S., Gupta, A., Levine, S., Hausman, K., Finn, C.: Gradient
  surgery for multi-task learning. In: Advances in Neural Information
  Processing Systems. vol.~33, pp. 5824--5836 (2020)

\bibitem{zala2023hierarchical}
Zala, A., Cho, J., Kottur, S., Chen, X., Oguz, B., Mehdad, Y., Bansal, M.:
  Hierarchical video-moment retrieval and step-captioning. In: CVPR. pp.
  23056--23065 (2023)

\bibitem{zeng2026promptemo}
Zeng, X., Yang, Y., Zeng, P., Yin, W., Liu, B., Wu, X., Wang, Y.: Promptemo:
  Learning emotion with bilateral textual prompts in multi-domain open-set
  scenarios. In: Proceedings of the AAAI Conference on Artificial Intelligence.
  vol.~40, pp. 12322--12330 (2026)

\bibitem{zhao2025weakly}
Zhao, C., Zhao, Z., Zhao, X.: Weakly-supervised video highlight detection by
  characteristic and commonality modeling. In: ICASSP. pp.~1--5 (2025)

\end{thebibliography}

\end{document}